%% file: Part_II_revised3.tex
\documentclass[final,3p,times,twocolumn]{elsarticle}

\usepackage{amsmath}
\usepackage{amssymb}
\usepackage{amsthm}
\usepackage{booktabs}
\usepackage{breqn}
\usepackage{cancel}
\usepackage{caption}
\usepackage{subcaption}
\usepackage{chemarrow}
\usepackage{chemformula}
\usepackage{chngcntr}
\usepackage{color}
\usepackage{colortbl}
\usepackage{cuted}
\usepackage{dblfloatfix}
\usepackage{enumitem}
\usepackage{epsfig}
\usepackage{epstopdf}
\usepackage{esint}
\usepackage{etoolbox}
\usepackage{geometry}
\usepackage{graphics}
\usepackage{graphicx}
\usepackage{harpoon}
\usepackage{hyperref}
\usepackage{latexsym}
\usepackage{lineno}
\usepackage{lipsum}
\usepackage{longtable}
\usepackage{lscape}
\usepackage{makecell,multirow}
\usepackage{mathrsfs}
\usepackage{mathtools}
\usepackage{mhchem}
\usepackage{multicol}
\usepackage{soul}
\usepackage{threeparttable}
\usepackage{tikz}
\usepackage[tight,nice]{units}
\usepackage{xcolor}

\setlist[itemize]{noitemsep, nolistsep}
\definecolor{LightCyan}{rgb}{0.88,1,1}
\definecolor{Gray}{gray}{0.85}
\hypersetup{
    colorlinks=false,
    linkcolor=black,
    filecolor=magenta,      
    urlcolor=black,
    }

\biboptions{sort&compress}

\makeatletter
\def\ps@pprintTitle{%
  \let\@oddhead\@empty
  \let\@evenhead\@empty
  \let\@oddfoot\@empty
  \let\@evenfoot\@oddfoot
}

\newcommand{\revone}[1]{\textcolor{black}{{#1}}}
\newcommand{\revtwo}[1]{\textcolor{black}{{#1}}}
\newcommand{\revall}[1]{ \textcolor{black}{{#1}}}
\newcommand{\revtwotwo}[1]{ \textcolor{black}{{#1}}}
\newcommand{\revtwothree}[1]{ \textcolor{black}{{#1}}}

\newrobustcmd*{\myVtriangle}[2]{\tikz{\filldraw[draw=#1,fill=#2] (0cm,0.2cm) --
(0.2cm,0.2cm) -- (0.1cm,0cm) -- (0cm,0.2cm);}}

\newrobustcmd*{\mythickVtriangle}[2]{\tikz{\filldraw[line width=0.3mm,draw=#1,fill=#2] (0cm,0.2cm) --
(0.2cm,0.2cm) -- (0.1cm,0cm) -- (0cm,0.2cm);}}

\newrobustcmd*{\mythickErrorVtriangle}[2]{\tikz{\filldraw[line width=0.3mm,draw=#1,fill=#2] (-0.05cm,0.05cm) --
(0.05cm,0.05cm) -- (0cm,-0.05cm) -- (-0.05cm,0.05cm);  \draw[draw=#1] (0.0cm, -0.12cm) -- (0.0cm, 0.12cm) ; \draw[draw=#1] (-0.06cm, 0.12cm) -- (0.06cm, 0.12cm); \draw[draw=#1] (-0.06cm, -0.12cm) -- (0.06cm, -0.12cm)    }}

\newrobustcmd*{\mytriangle}[2]{\tikz{\filldraw[draw=#1,fill=#2] (0.0cm,0.0cm) --
(0.2cm,0cm) -- (0.1cm,0.2cm) -- (0cm,0cm);}}

\newrobustcmd*{\mysquare}[2]{\tikz{\draw[draw=#1,fill=#2] (0cm,0cm)
rectangle (0.2cm,0.2cm)}}

\newrobustcmd*{\mythicktriangle}[2]{\tikz{\filldraw[line width=0.3mm,draw=#1,fill=#2] (0.0cm,0cm) --
(0.2cm,0cm) -- (0.1cm,0.2cm) -- (0.0cm,0cm);}}

\newrobustcmd*{\mythicksquare}[2]{\tikz{\draw[line width=0.3mm,draw=#1,fill=#2] (0cm,0cm)
rectangle (0.2cm,0.2cm)}}

\newrobustcmd*{\mybarredtriangle}[2]{\tikz{\draw[draw=#1,fill=#2] (0,0) --
(0.2cm,0) -- (0.1cm,0.2cm) -- (0cm,0cm); \draw[draw=#1] (-0.1cm, 0.07cm) -- (0.3cm, 0.07cm)}}

\newrobustcmd*{\mythickbarredtriangle}[2]{\tikz{\draw[line width=0.3mm,draw=#1,fill=#2] (0,0) --
(0.2cm,0) -- (0.1cm,0.2cm) -- (0cm,0cm); \draw[draw=#1] (-0.1cm, 0.07cm) -- (0.3cm, 0.07cm)}}

\newrobustcmd*{\mybarredsquare}[2]{\tikz{\draw[draw=#1,fill=#2] (0,0)
rectangle (0.2cm,0.2cm); \draw[draw=#1] (-0.1cm, 0.1cm) -- (0.3cm, 0.1cm)}}

\newrobustcmd*{\mythickbarredsquare}[2]{\tikz{\draw[line width=0.3mm,draw=#1,fill=#2] (0,0)
rectangle (0.2cm,0.2cm); \draw[draw=#1] (-0.1cm, 0.1cm) -- (0.3cm, 0.1cm)}}

\newrobustcmd*{\mybarredcircle}[2]{\tikz{\draw[draw=#1,fill=#2] (0,0)
circle (0.1cm); \draw[draw=#1] (-0.2cm, 0.0cm) -- (0.2cm, 0.0cm)}}

\newrobustcmd*{\mythickbarredcircle}[2]{\tikz{\draw[line width=0.3mm,draw=#1,fill=#2] (0,0)
circle (0.1cm); \draw[draw=#1] (-0.2cm, 0.0cm) -- (0.2cm, 0.0cm)}}

\newrobustcmd*{\mythickErrorcircle}[2]{\tikz{\draw[line width=0.3mm,draw=#1,fill=#2] (0,0)
circle (0.06cm); \draw[draw=#1] (0.0cm, -0.12cm) -- (0.0cm, 0.12cm) ;   \draw[draw=#1] (-0.06cm, 0.12cm) -- (0.06cm, 0.12cm); \draw[draw=#1] (-0.06cm, -0.12cm) -- (0.06cm, -0.12cm)    }}

\newrobustcmd*{\mydashedline}[1]{\tikz{\draw[draw=#1] (-0.2cm, 0.2cm) -- (-0.1cm, 0.2cm); \draw[draw=#1] (-0.0cm, 0.2cm) -- (0.1cm, 0.2cm)}}

\newrobustcmd*{\mythickcross}[1]{\tikz{\draw[line width=0.3mm,draw=#1] (0,0) --
(0.2cm,0); \draw[line width=0.3mm,draw=#1] (0.1cm,-0.1cm) -- (0.1cm,0.1cm);}}

\newrobustcmd*{\mybarredcross}[1]{\tikz{\draw[line width=0.3mm,draw=#1] (0,0) --
(0.2cm,0); \draw[line width=0.3mm,draw=#1] (0.1cm,-0.1cm) -- (0.1cm,0.1cm); \draw[draw=#1] (-0.1cm,0) -- (0.3cm,0);}}

\newrobustcmd*{\myline}[1]{\tikz{\draw[draw=#1] (-0.15cm, 0.1cm) -- (0.15cm, 0.1cm);\draw[line width=0.3mm,draw=#1] (-0.0cm, 0.0cm);}}

\newrobustcmd*{\mythickline}[1]{\tikz{\draw[line width=0.3mm,draw=#1] (-0.15cm, 0.1cm) -- (0.15cm, 0.1cm);\draw[line width=0.3mm,draw=#1] (-0.0cm, 0.0cm);}}

\newrobustcmd*{\mythickdashedline}[1]{\tikz{\draw[line width=0.3mm,draw=#1] (-0.2, 0.1cm) -- (-0.1cm, 0.1cm); \draw[line width=0.3mm,draw=#1] (-0.0cm, 0.1cm) -- (0.1cm, 0.1cm); \draw[line width=0.3mm,draw=#1] (-0.0cm, 0.0cm);}}

\newrobustcmd*{\mythickdasheddottedline}[1]{\tikz{\draw[line width=0.3mm,draw=#1] (-0.22, 0.1cm) -- (-0.13cm, 0.1cm); \draw[line width=0.3mm,draw=#1] (-0.085cm, 0.1cm) -- (-0.055cm, 0.1cm); \draw[line width=0.3mm,draw=#1] (-0.01cm, 0.1cm) -- (0.08cm, 0.1cm); \draw[line width=0.3mm,draw=#1] (-0.0cm, 0.0cm);}}

\newrobustcmd*{\mycircle}[2]{\tikz{\draw[draw=#1,fill=#2] (0,0)
circle (0.1cm);}}

\newrobustcmd*{\mythickcircle}[2]{\tikz{\draw[line width=0.3mm,draw=#1,fill=#2] (0,0)
circle (0.1cm);}}

\newrobustcmd*{\mydot}[1]{\tikz{\draw[line width=0.3mm,draw=#1] (0,0)
circle (0.025cm);}}

\journal{TBD}

\begin{document}

\begin{frontmatter}

\title{PINN surrogate of Li-ion battery models for parameter inference. \\Part II: Regularization and application of the pseudo-2D model}

\author[NREL_CompSci]{Malik Hassanaly\corref{cor}}
	\cortext[cor]{Corresponding author. Tel: (303) 275-4739.}
		\ead{malik.hassanaly@nrel.gov}
\author[NREL_ECaSS]{Peter J. Weddle}
\author[NREL_CompSci]{Ryan N. King}
\author[NAU]{Subhayan De}
\author[UC_Boulder]{Alireza Doostan}
\author[NREL_ECaSS]{\\Corey R. Randall}
\author[INL]{Eric J. Dufek}
\author[NREL_ECaSS]{Andrew M. Colclasure}
\author[NREL_ECaSS]{Kandler Smith}

\address[NREL_CompSci]{Computational Science Center, National Renewable Energy Laboratory (NREL), Golden, CO 80401}
\address[NREL_ECaSS]{Energy Conversion and Storage Systems Center, National Renewable Energy Laboratory, Golden, CO 80401}
\address[UC_Boulder]{Aerospace Mechanics Research Center, University of Colorado, Boulder, CO 80303}
\address[NAU]{Mechanical Engineering Department, Northern Arizona University, Flagstaff, AZ 86011}
\address[INL]{Energy Storage and Electric Transportation Department, Idaho National Laboratory (INL), Idaho Falls, ID 83415}

\begin{abstract} 
Bayesian parameter inference is useful to improve Li-ion battery diagnostics and can help formulate battery aging models.  However, it is computationally intensive and cannot be easily repeated for multiple cycles, multiple operating conditions, or multiple replicate cells.  To reduce the computational cost of Bayesian calibration, numerical solvers for physics-based models can be replaced with faster surrogates.  A physics-informed neural network (PINN) is developed as a surrogate for the pseudo-2D (P2D) battery model calibration.  For the P2D surrogate, additional training regularization was needed as compared to the PINN single-particle model (SPM) developed in Part~I.  Both the PINN SPM and P2D surrogate models are exercised for parameter inference and compared to data obtained from a direct numerical solution of the governing equations.  A parameter inference study highlights the ability to use these PINNs to calibrate scaling parameters for the cathode \ce{Li} diffusion and the anode exchange current density.  By realizing computational speed-ups of $\approx$2250x for the P2D model, as compared to using standard integrating methods, the PINN surrogates enable rapid state-of-health diagnostics. In the low-data availability scenario, the testing error was estimated to $\approx$2mV for the SPM surrogate and $\approx$10mV for the P2D surrogate which could be mitigated with additional data.
\end{abstract}



\end{frontmatter}

\section{Introduction}
\label{sect:Introduction}

As batteries are pivotal to today's economy, developing tools that accurately diagnose and forecast battery state-of-health is essential~\cite{RENIERSHOWEY21} to appropriately manage \ce{Li}-ion batteries degradation~\cite{EOPKPHGACYLPDTMRWPWO21}. In the present work, the battery's internal parameters are determined from the voltage response during a discharge cycle, through Bayesian calibration.  Bayesian calibration for parameter inference is computationally expensive because the underlying physics model that maps the internal parameters to the observed voltage response must be run many times~\cite{KKCC23}.  Here, a physics-informed neural network (PINN) is trained to replace the more computationally expensive pseudo-2D (P2D) battery model.  

In Part~I \revone{\cite{hassanaly2023pinn1}}, a framework is discussed to develop a PINN for the single-particle model (SPM).  A companion repository  (\hyperlink{https://github.com/NREL/PINNSTRIPES}{https://github.com/NREL/PINNSTRIPES}) is provided.  Here, the framework is extended to capture the P2D model physics.  The P2D model is a standard physics-based model in the Li-ion battery community that captures heterogeneous electrode utilization during high-rate cycling~\cite{FDN94,ADW99,SGRW06,CDTPJS19,WKCYGCSGTD23}. Using surrogates of higher fidelity models is critical to mitigate modeling error and better capture the effect of additional aging properties. After training, a P2D PINN surrogate model is deployed to accelerate Bayesian calibration and identify the internal battery parameters using a discharge (2~C) voltage response.  Similar to Part~I \revone{\cite{hassanaly2023pinn1}}, an emphasis is placed on the training procedure that can successfully train the neural net using only the physics-informed loss, which mimics a typical data-poor parameter inference in high dimensions. 

\subsection{Prior works}
\label{PriorWorks}

Li-ion batteries can age in different ways ~\cite{VMN10,GLBBJHLD22,TWYCCFLPKAUCBDSCWWY22,TYFCLWBCDWTESADQDTJ22,MLSS21,PTCSTDDTJTW21,DABCCCGKKMRRSTUW22,BRMBH17, VNWVMBWWVH05} making aging forecast and diagnostics challenging. State-of-health diagnostics can be divided into 1) experimental observations and 2) model inferences, with significant overlap between these approaches. For non-intrusive experimental measurements~\cite{MIGUELPLETT21,LTP22}, a significant amount of work focused on the evolving electrochemical signatures with aging~\cite{THEBCSTMD21,FBSYHJSSD14,YCLTYWJTWRB22,FBSYHJSSD14,MLSS21,SKSS16,DB22}. Model-centric techniques that determine a battery's internal state rely on electrochemical signals~\cite{DB22,KKCC23,SMALYSKLSSH21,SMALYSKLSSH21} or externally measured pressure changes~\cite{SKSS16} to infer internal aging dynamics.  These techniques typically determine high-level aging ``modes'' such as loss-of-active-material (LAM), loss-of-lithium-inventory (LLI), resistance growth, and/or capacity fade~\cite{WKCYGCSGTD23,LSSLL20,CSAD22,DA22}.  

An alternative approach is to map the measured aged response to internal parameter dynamics~\cite{NJI13,PB13,SMPSS21}.  Parameter inference necessarily requires a proposed model that translates property variations to changes in cell performance. Generally, inverse parameter inference is accomplished by either finding an optimal parameter set that best explains experimental observations~\cite{MSUKUYU23,ASKKKEJL22,LDCJRJS22}, or by finding a distribution of candidate parameter values (e.g., Bayesian parameter calibration) that describe the measured responses~\cite{GCHJS22,AMAH20,KKCC23}.  The former approach is more computationally efficient and has been used to calibrate up to 44 internal P2D parameters~\cite{ASKKKEJL22,ZLWZLM13,reddy2019accelerating}. The latter approach can be extremely computationally expensive and requires a fast underlying model, and has recently been to calibrate 15 parameters of the P2D model~\cite{KKCC23}. 

Importantly, when determining internal parameters, three sources of uncertainty coexist \revtwothree{as noted in \cite{guo2022review}}. First, the available experimental observation may be insufficient to accurately infer the battery's internal parameters~\cite{HMD15,CD17}. This is referred to as coarse-graining uncertainty, which is typical of partially observed systems~\cite{hassanaly2022adversarial,rybchuk2023ensemble,BFSV23,ASKKKEJL22}. Coarse-graining uncertainty may affect each parameter differently, resulting in varying levels of identifiability~\cite{li2022data, bizeray2018identifiability,MIGUELPLETT21}. Second, experimental observations may be subject to noise (e.g., Maccor Series 4000 has a voltage accuracy of $\approx$3~mV)\footnote{http://www.maccor.com/Products/Series4000.aspx}. Third, the physics-based models used to interpret the parameters' influences on observed responses may inaccurately depict the real system. One way to include the aforementioned uncertainties in the inferred parameter uncertainties is \revtwothree{to use Bayesian calibration. Bayesian treatment of uncertainties has been long recognized as a necessary component of inverse modeling tasks in energy storage applications. It was applied in the form of particle filtering for estimating internal parameters \cite{bi2020online} or state-of-charge and remaining useful life \cite{saha2007integrated,liao2016hybrid}, and in the form of Kalman filtering \cite{li2020electrochemical, zheng2023state} for internal state estimation. A basic description of the Bayesian calibration procedure adopted in this work is provided in Section~\ref{sect:BayesianCalibration}.}

The present work uses a PINN surrogate model to improve the computational tractability of parameter inference using Bayesian calibration schemes.  After developing a PINN that reliably converts input parameters to observed voltage responses, Bayesian calibration can be used to ask: ``what parameters are most likely to produce the observed voltage response?''. By developing this tool, internal property dynamics (and confidence intervals on these parameters) during aging can be extracted from voltage/current responses. \revtwothree{In this study, parameter inference is done using constant-current conditions, which is commonly used when studying battery degradation~\cite{KYCTD22,WKCYGCSGTD23,GSSEST23,SSKLCR17,HMLLLZMGCGBCSD19,WHOP13}. For such studies, rapid and uncertainty-aware parameter identification allows for diagnosing the degradation mechanism at stake, as well as the rate at which internal battery properties degrade. Extensions to dynamic currents can be achieved within the same framework at the expense of potentially needing to use larger surrogates (see Part~I \cite{hassanaly2023pinn1}, Appendix B). Dynamic current extensions can also be achieved by using a neural operator approach to encode the time-dependent current load \cite{zheng2023inferring}, or a transfer learning approach to leverage the charge dynamics already encoded at constant current \cite{bhattacharjee2021estimating}.} Notably, PINN strategies~\cite{raissi2019physics} have been used previously in the literature as redox-flow battery model surrogates~\cite{he2022physics,CHEN2023233548}, and for low-fidelity Li-ion battery models, such as the SPM~\cite{singh2023hybrid} or the Verhulst model~\cite{wen2023fusing}. \revtwothree{Finally, physics-informed neural operator strategies have also been employed to approximate the extended SPM model \cite{zheng2023state}. To the authors' knowledge, the present work is the first implementation of a PINN that predicts P2D model solutions, and the first demonstration of the successful training of a P2D PINN surrogate in a data-poor environment. The application of a physics-informed surrogate for parameter inference was conducted before in Ref.~\cite{zheng2023inferring}. However, the uncertainty quantification strategy in Ref.~\cite{zheng2023inferring} used a heuristic approach by repeating deterministic parameter inference. In this work, we utilize a key advantage of physics-informed methods: given their ability to be accurate over a large input domain (Sec.~\ref{sec:parametericPinn}), they are best suited for performing sample-intensive tasks such as Bayesian parameter inference (Sec.~\ref{sec:cal}).}

The \revtwothree{novel contributions} of this manuscript are as follows
\begin{itemize}
    \item We demonstrate for the first time the ability of PINN to predict P2D model solutions in a data-poor environment for the first time granted training regularization that we derive. (Sec.~\ref{sec:p2dSurr}).
    \item We show that multi-fidelity hierarchical training is beneficial for the PINN P2D surrogate and for parametric PINNs  (Sec.~\ref{sec:parametericPinn}).
    \item We assess the effect of physics and data loss in a data-poor environment (Sec.~\ref{sec:parametericPinn}).
    \item We demonstrate the computational benefits of PINNs for \revtwothree{Bayesian parameter inference} (Sec.~\ref{sec:cal}). 
\end{itemize}


\subsection{Bayesian calibration}
\label{sect:BayesianCalibration}

Bayesian calibration in the context of parameter inference for \ce{Li}-ion batteries is typically an expensive calculation with increased cost when considering every cycle and every observed battery response~\cite{bills2023massively,aitio2020bayesian,BFSV23,KKCC23}. There are at least two strategies that reduce the computational cost of Bayesian calibration without losing cycle- or cell-specific fidelity.  The first strategy is to use insights from data to limit the number of inferred parameters (based on prior knowledge of battery degradation).  However, there is evidence that reducing the number of parameters \textit{a priori} can significantly affect the posterior information sought about the parameters and the resulting model~\cite{ramadesigan2011parameter}.  In other words, reducing the Bayesian calibration to fewer parameters can result in non-unique parameter sets that equally explain the experimental data.  A second approach is to reduce the likelihood function $p_{\rm like}$ evaluation cost.  In this manuscript, the main objective is to provide a framework for constructing computationally inexpensive \ce{Li}-ion battery surrogate models to accelerate the likelihood function evaluation. 

In Bayesian calibration, one seeks the posterior probability of battery parameters $p_{\rm post}(\boldsymbol{p}|\boldsymbol{d})$, where $\boldsymbol{p}$ represents the internal parameters, and $\boldsymbol{d}$ the experimental observations. The posterior probability $p_{\rm post}$ is the probability of a particular parameter value $\boldsymbol{p}$ given observation $\boldsymbol{d}$.  The posterior probability $p_{\rm post}$ is obtained by applying the Bayes theorem 
\begin{equation}
    p_{\rm post}(\boldsymbol{p}|\boldsymbol{d}) \propto p_{\rm prior}(\boldsymbol{p}) p_{\rm like} (\boldsymbol{d} | \boldsymbol{p}), 
\end{equation}
where $p_{\rm prior}(\boldsymbol{p})$ is the prior probability that encodes prior knowledge about the parameters (e.g., maximum and minimum values) and $p_{\rm like}(\boldsymbol{d} | \boldsymbol{p})$ is the likelihood function that characterizes how likely a parameter set $\boldsymbol{p}$ is to explain the observed data $\boldsymbol{d}$. In other words, the likelihood function gives the probability that a particular parameter set $\boldsymbol{p}$ explains the experimental observation. To compute the likelihood function, one needs to choose a parameter set $\boldsymbol{p}$, use this parameter set to make a prediction of the observation, and measure the discrepancy of the predicted observation to the realized observation $\boldsymbol{d}$. In Bayesian calibration, coarse-graining uncertainty is automatically captured because multiple parameter samples can be proposed to explain the same observational data, while the experimental noise and model inaccuracies can be lumped into the uncertainty of the likelihood function $p_{\rm like}$~\cite{hassanaly2021surface,braman2013bayesian}. 

The primary downside of Bayesian calibration is that it requires many model evaluations to appropriately delineate the support of the posterior distribution. At minimum a model evaluation at each parameter value where the posterior is sought in $\boldsymbol{p}$-space (also shown in Appendix~\ref{sect:SPMBrute}). In high dimensions, a common practice is to draw samples from the posterior distribution via Markov chain Monte Carlo (MCMC) methods~\cite{tierney1994markov}. In MCMC, a sequence of samples that span the $\boldsymbol{p}$-space is constructed such that the samples converge to independent samples of the posterior distribution. At every step of the sequence, an MCMC procedure requires an expensive evaluation of the likelihood function, by interrogating the numerical model for every available observational data point. The cost of MCMC is also plagued by a ``warm-up'' time needed before convergence to the posterior is achieved, during which discarded samples are also generated by evaluating the likelihood function~\cite{roberts1994simple,gelman1997weak}.  

Consistent with other analyses~\cite{braman2013bayesian}, the likelihood function is assumed to be a multivariate normal distribution given by
\begin{equation}
\label{eq:like}
\begin{split}
    p_{\rm like}(\boldsymbol{d} | \boldsymbol{p}) & = \frac{1}{(2\pi \sigma^2)^{N_{\rm d}/2}} \\ & \times \ \exp \left[ - \frac{1}{2 \sigma^2} \sum_{i=1}^{N_{\rm d}} (\boldsymbol{d} - \boldsymbol{d}_{\rm pred}(\boldsymbol{p}))^2   \right], 
\end{split}
\end{equation}
where $N_{\rm d}$ is the number of observations, $\sigma$ is the standard deviation of the multivariate normal distribution (where uncertainty is assumed to be the same for all observations), and $\boldsymbol{d}_{\rm pred} (\boldsymbol{p})$ is the prediction of observations that would be obtained if parameter set $\boldsymbol{p}$ was chosen. The computation of $\boldsymbol{d}_{\rm pred}$ is the expensive step that normally requires evaluating a physics-based model.  Using a pure optimization method -- i.e., minimizing $\sum_{i=1}^{N_d} (\boldsymbol{d} - \boldsymbol{d_{pred}}(\boldsymbol{p}))^2$ -- implicitly assumes that any mismatch between the predicted and observed data is due to a misspecification of $\boldsymbol{p}$. In reality, even if $\boldsymbol{p}$ is exactly identified, $\boldsymbol{d}$ might not exactly match $\boldsymbol{d}_{\rm pred}$ because of experimental or modeling errors that provide the mapping between $\boldsymbol{p}$ and $\boldsymbol{d}_{\rm pred}$. While both errors may be lumped into $\sigma$, a high uncertainty reduces the confidence in the inferred parameters, as shown in Sec.~\ref{sec:cal}. Mitigating model errors is possible by adopting a higher fidelity model for the \ce{Li}-ion battery system (e.g., using the P2D model as opposed to the SPM model).



\begin{figure*}[th!]
    \centering
    \includegraphics[width=0.9\textwidth]{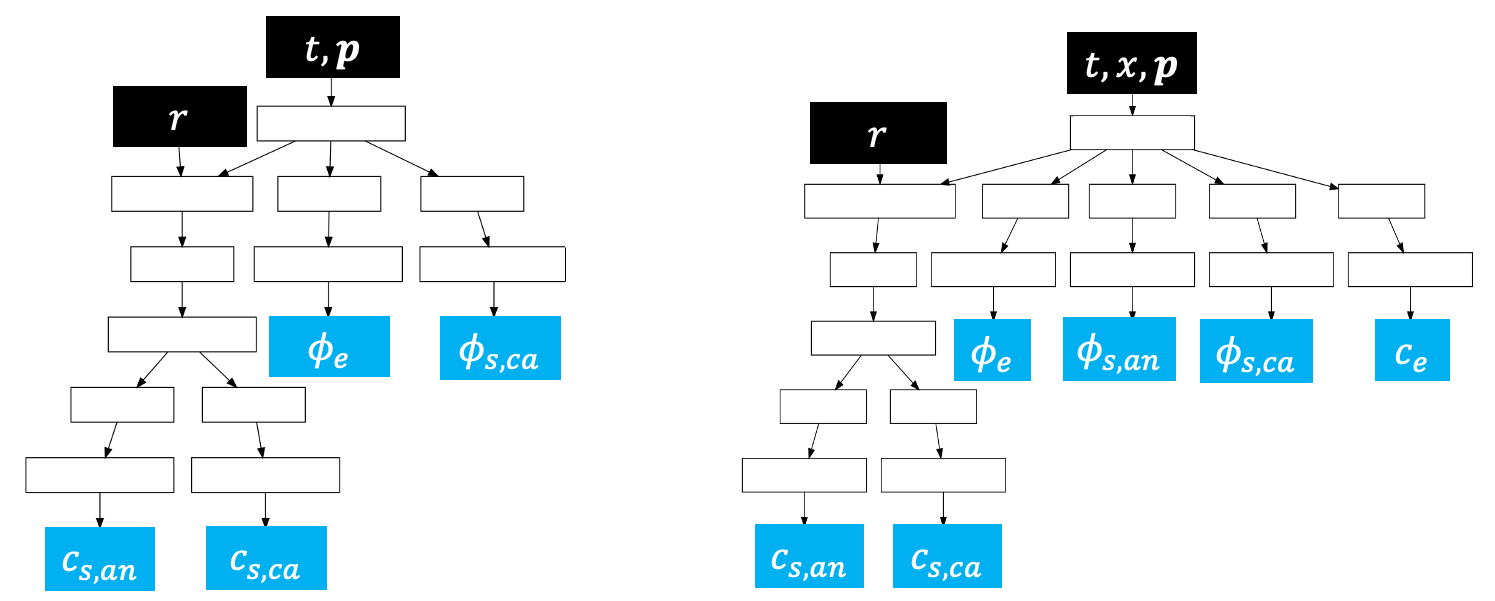}
    \caption{PINN architecture schematic used to enforce spatiotemporal and parametric dependencies of the state variables for the SPM (left) and the P2D model (right). \revone{White rectangles denote blocks of hidden layers that could be of any type}.}
    \label{fig:arch}
\end{figure*}

\section{Pseudo-2D battery model}
\label{sect:P2DModel}

The P2D battery model is a physics-based model that captures kinetic and transport resistances and heterogeneous utilization across a cell electrode assembly~\cite{DFN93,DFN94b,FDN94,DFN95,NT04,CK10,KSLSP11,RN97,W19,CTJTDPBDFEDS20,GSW11,SB17}.  The P2D model resolves the \ce{Li} species and charge transport within both composite electrodes, the separator, and the percolating electrolyte.  The model has two coordinate directions:  the primary direction is normal to the current collectors (also referred to as through-plane), and the secondary direction radially points outwards through the representative spherical electrode particles.  

In the P2D model, there are four state variables, including the liquid-phase Li-ion concentration within the electrolyte $c_{\rm e}$, the liquid electrolyte potential $\phi_{\rm e}$, the solid-phase electrode potential $\phi_{{\rm s},j}$, and the solid-phase Li concentration $c_{{\rm s},j}$.  The subscript $j$ indicates the domain (i.e., anode, separator, or cathode).  All state variables are resolved in the primary direction normal to the current collector (i.e., the $x$-direction) aside from the solid-phase Li concentration $c_{{\rm s},j}$, which is resolved in the secondary radial direction $r$.  The P2D governing equations are provided in Appendix~\ref{sect:P2DModel_Appendix}.

The coupled system of differential-algebraic equations used in the P2D model is significantly more complicated and costly than the decoupled, ordinary, differential equations used in the SPM (cf., Part~I \revone{\cite{hassanaly2023pinn1}}).  Additionally, additional material/transport/architecture parameters are introduced and are useful for studying heterogeneous (through-plane) electrode utilization, and electrolyte resistance/depletion that are non-negligible in high-rate ($\geq$2~C) demands. For the present study, the PINN P2D surrogate is trained using parameters from Colclasure et al.~\cite{CDTPJS19}. Some of the parameters are later modified in the manuscript to study the PINN performance in determining ``unknown'' parameter sets (Section~\ref{sec:application}).


\section{Method: PINN surrogate for the P2D model}
\label{sec:p2dSurr}

The P2D PINN surrogate construction follows similar design guidelines to those developed in Part~I \revone{\cite{hassanaly2023pinn1}}. Compared to the SPM, the P2D model involves an additional longitudinal spatial variable $x$ (cf., Fig.~\ref{fig:arch}). The additional variables used in the P2D model, as compared to the SPM, are the potential in the negative electrode $\phi_{\rm s,an}$ and the concentration of \ce{Li}-ions in the electrolyte $c_{\rm e}$.  Additionally, the model parameters $\boldsymbol{p}$ are shown as inputs to both the SPM and the P2D model in Fig.~\ref{fig:arch}. Including the parameters as inputs (which was not used in Part~I \revone{\cite{hassanaly2023pinn1}}) is needed to apply the surrogate models to Bayesian calibration (i.e., the neural networks need to encode the dependence of the solutions with respect to the parameters $\boldsymbol{p}$ being inferred). The parameters $\boldsymbol{p}$ can be treated similarly to the spatiotemporal variables -- $t$ for the SPM, and $t$ and $x$ for the P2D model -- that affect the prediction of all the state variables. 

\subsection{Initial conditions enforcement}
\label{sec:InitCond}

To prevent predicting $\phi_{\rm s, an}$ up to a constant shift as noted by Chen et al.~\cite{CHEN2023233548}, the Dirichlet boundary condition for $\phi_{\rm s,an}$ at $x=0$ is strictly enforced similarly to the strict initial conditions enforcement described in Part~I \revone{\cite{hassanaly2023pinn1}},
\begin{equation}
\begin{split}
    \label{eq:spacdist}
    \phi_{\rm s,an}(x) &= F(t) \left[ \widetilde{\phi}_{\rm s,an}(x, t) F(x) + \phi_{\rm s,an}(x=0) \right] \\ &+ \phi_{\rm s,an}(x, t=0).
\end{split}
\end{equation}
Here, \revtwo{$\widetilde{\phi}_{\rm s,an}(t,x)$ is the raw output of the neural net, $F(t) = 1 - \exp(-t/\tau)$}, $\tau$ is a timescale over which the initial condition is enforced (assumed to be 1~s), $F(x) = x / L_{\rm an}$ consistently with Sun et al.~\cite{sun2020surrogate}, $\phi_{\rm s,an}(x=0)$ is the strictly enforced boundary condition, and $\phi_{\rm s,an}(x, t=0)$ is the strictly enforced initial condition (if any, as discussed below). 


\begin{figure*}
     \centering
        
        
    \includegraphics[width=6.497in]{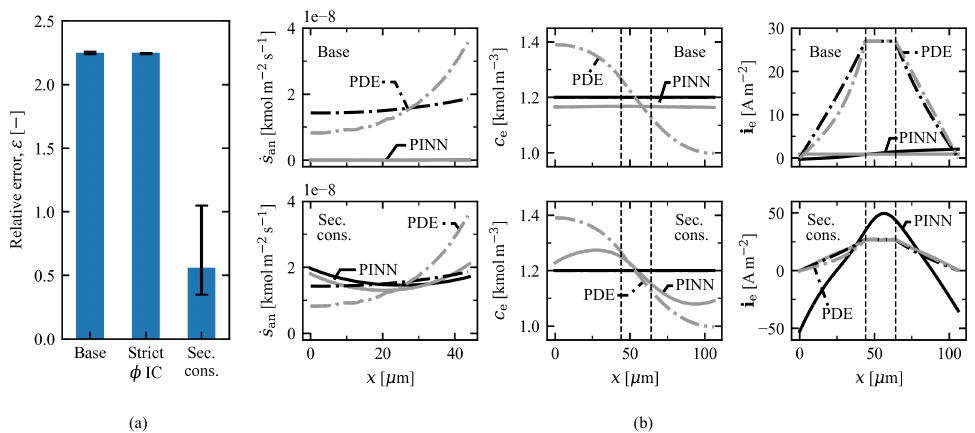}
    \caption{\revtwotwo{(a) Average PINN error over the training realizations (bar height) that illustrates the effect of secondary conservation regularization and strict enforcement of initial potential conditions. The error bar denotes the 95\% percentile variability observed for all the realizations. (b) Prediction of Faradaic current in the anode (left), electrolyte \ce{Li}-ion concentration (middle) and current (right), without secondary regularization (top) and with secondary regularization (bottom). Variables are shown after 0~s (black) and 400~s (gray). The profiles obtained from the PDE are shown as dashed-dotted lines with the same color coding. The electrode/separator boundaries are shown for the electrolyte \ce{Li}-ion concentration and current (vertical black dashed line).}}
    \label{fig:sec_cons}
\end{figure*}

Strict initial conditions enforcement is more difficult for the P2D model as compared to the SPM due to the potential in the negative electrode $\phi_{\rm s,an}$, in the positive electrode $\phi_{\rm s,ca}$, and in the electrolyte $\phi_{\rm e}$ having spatial dependence. Unlike the \ce{Li} concentrations, the potential spatial distributions algebraically depend on the prescribed initial concentrations. In the case of the SPM, the potentials were not spatially dependent and their initial concentrations could be precomputed, and strictly enforced during training. In the case of the P2D model, prescribing the initial conditions of the potentials requires computing the entire initial spatial distribution. In Eq.~\ref{eq:spacdist}, this is emphasized by preserving the spatial dependence of the initial condition $\phi_{\rm s,an}(x, t=0)$. While prescribing the analytical solution of the initial potential values is possible, the analytical solution would need to encode the dependence of the initial conditions on the internal parameters $\boldsymbol{p}$. 

To resolve the initial potentials in the P2D model, two methods are adopted and compared hereafter: 1) the initial conditions are not strictly enforced but are captured by virtue of minimizing the residuals near $t=0$. In this case, the potentials in Eq.~\ref{eq:spacdist} only use an initial guess for the initial condition $\xi_{0}$ where $\xi$ denotes a state variable, and the distance function $F(t)=1$, thereby allowing to adjust the initial guess; 2) the initial conditions are computed and encoded by a neural net by training another PINN over the time interval [$0$, $0$]~s. The advantage of the second method is that the PINN in charge of learning the initial conditions will naturally encode the dependence of the initial conditions with the internal parameters being varied. The drawback of the second method is that an extra PINN evaluation is needed when predicting the state variables. In practice, the PINN in charge of the initial conditions learning can use a smaller network since it does not aim to capture any temporal variability. When used here, the neural network that encodes the initial conditions uses 12 neurons per layer (as opposed to 20 for the spatiotemporal PINN) and only one gradient pathology block (as opposed to 3 for the spatiotemporal PINN). In turn, the added computing overhead due to the extra PINN evaluation is limited to about 5\% of the total training time for the cases investigated here. Both methods are evaluated in Sec.~\ref{sec:fail} and Sec.~\ref{sec:SecondaryConservation}.

\subsection{Training failure mode}
\label{sec:fail}

The PINN design guiding principles identified for the SPM in Part~I \revone{\cite{hassanaly2023pinn1}} are reused here for the P2D model. Specifically, a merged architecture with gradient pathology blocks \cite{wang2021understanding} is used. For now, a linearized Butler--Volmer formulation is implemented and no hierarchical training is used. This training method is referred to as \textit{Base}. Given the added spatial variable $x$, as compared to the SPM, the number of collocation points is twice that of the SPM (described in Part~I \revone{\cite{hassanaly2023pinn1}}). The total number of epochs (number of times the entire training set is shown to the network) is also doubled. The merged neural net architecture contains two layers in the network trunk (compared to one for the SPM) and the three gradient pathology blocks in the network branches. The number of neurons per layer is held at 20 and the activation function is a hyperbolic tangent. These choices are held throughout the manuscript. Compared to the SPM, the learning rate is decreased by a factor of 10 but follows the same schedule. Similar to Part~I \revone{\cite{hassanaly2023pinn1}}, the accuracy of the model is evaluated with a scaled mean absolute error $\varepsilon$ defined as

\begin{equation}
    \label{eq:errP2D}
    \varepsilon = \sum_{\xi \in \{c_{\rm s,an}, c_{\rm s,ca}, c_{\rm e}, \phi_{\rm e}, \phi_{\rm s,an}, \phi_{\rm s,ca}\}}  \frac{1}{N_{\xi}} \sum_{i \in [1, N_{\xi}]} \left| \frac{\xi_{\rm PINN, i} - \xi_{\rm PDE, i}}{\xi_{\rm PDE, i} } \right|,
\end{equation}
where $N_{\xi}$ is the number of points over which the error is computed for each state variable $\xi$.

Figure~\ref{fig:sec_cons}a illustrates the \textit{Base} case error $\varepsilon$. Although the error definition contains two additional variables, which artificially increases the value of $\varepsilon$, the error observed is about 20 times higher as compared to errors from the SPM in Part~I \revone{\cite{hassanaly2023pinn1}}. The results are also consistently inaccurate. An example of the predicted state variables is shown in Fig.~\ref{fig:sec_cons}b (top). Compared to the PDE solution, the surface net rates of progress for Li-ion production $\dot{s}_{\rm an}$ (left) and $\dot{s}_{\rm ca}$ (not shown for the sake of brevity) are severely under-predicted (see Appendix~\ref{sect:P2DModel_Appendix} for computing $\dot{s}_j$). This is a key complication as compared to the SPM model where the surface net rates of progress is known and assumed to be uniform and constant throughout time. Since $\dot{s}_j$ is tightly coupled to $c_{{\rm s},j}$ via the particle surface concentrations (Eq.~\ref{eqn:ParticleBC}), the mismatch for $\dot{s}_j$ pushes the solid \ce{Li} concentrations (not shown for the sake of brevity) to remain near their initial value. The electrolyte \ce{Li}-ion concentration $c_{\rm e}$ (middle) also remains near its initial value. Notably, $c_{\rm e}$ slowly decreases over time, which clearly violates the conservation of salt in the system. The flat profile of $c_{\rm e}$ is also explained by the fact that the current densities are nearly zero in both the anode and the cathode. Finally, the electrolyte current density $\mathbf{i}_{\rm e}$ is also consistently under-predicted since the electrolyte potential $\phi_{\rm e}$ is predicted to be uniform (not shown for the sake of brevity). Note that the under-predicted current also clearly violates the conservation of charge in the separator. Overall, the PINN attempts to push the output towards a trivial solution, which is that of the initial profile with near-zero gradient boundary conditions for the spherical particles. In the case of the SPM, this issue was resolved by sufficiently weighting the solid-phase \ce{Li} concentration gradient constraint at the particle surface. In the case of the P2D model, this strategy is not viable as the particle surface boundary is set by $\dot{s}_j$, which is itself under-predicted. 

As a first attempt to avoid the trivial failure mode, the authors tried to strictly enforce the potential initial conditions, which inherently strictly enforces the initial conditions for $\dot{s}_j$. This model is referred to as \textit{Strict $\phi$ IC}. Figure~\ref{fig:sec_cons}a illustrates the error of this approach compared to the \textit{Base} case. Unfortunately, the error is not reduced by this approach.  Further inspection of the predicted variables (not shown here) suggests that the same trivial failure mode occurs in the \textit{Strict $\phi$ IC} as in the \textit{Base} case. 

\subsection{Secondary conservation}
\label{sec:SecondaryConservation}

To address the PINN P2D surrogate trivial failure mode, it is proposed here to use secondary conservation constraint equations. In other numerical simulation applications, secondary conservation commonly refers to conservation that is not directly enforced, but is expected as a by-product of solving the governing equations. For instance, while entropy conservation~\cite{gouasmi2022entropy} or kinetic energy conservation~\cite{morinishi2010skew, hassanaly2018minimally} are not directly enforced in computational fluid dynamics, they can be obtained through a strategic discretization. However, solving secondary conservation equations in addition to the primary conservation equations and reconciling the solution obtained can be difficult with traditional numerical integrators. In the case of a PINN, there is no need to constrain the state variables only once and secondary constraints can be added as a physics-informed regularization. This strategy echos the gPINN method~\cite{yu2022gradient} that constrains the residual gradient as well as the residuals themselves. Unlike the generic gPINN approach, the objective is to derive specific constraints that address the training failure mode observed (i.e., a zero current density prediction). 

In the present approach, four additional constraint equations are added to the P2D model.  Importantly, the additional constraints do not add ``new physics'' to the P2D model but rather enforce a known consequence of the governing equations described in Appendix~\ref{sect:P2DModel_Appendix}.

First, conservation of charge requires that for all $x$ spatial locations the current is either carried in the electrolyte phase or in the solid-phase, which can be stated mathematically as 
\begin{equation}
    \nabla_x\cdot\left(\mathbf{i}_{\rm e} -\sigma^{\rm eff}_{{\rm s},j}\nabla_x\phi_{{\rm s},j}\right) = 0,
\end{equation}
where $\nabla_x$ indicates that the operator is in the $x$-direction, $\mathbf{i}_{\rm e}$ is the electrolyte current density, and $\sigma^{\rm eff}_{{\rm s},j}$ is the effective solid-phase conductivity.  Alternatively, propagating the cathode-current collector boundary condition results in 
\begin{equation}
    \mathbf{n}_x\cdot\left(\mathbf{i}_{\rm e} -\sigma^{\rm eff}_{{\rm s},j}\nabla_x\phi_{{\rm s},j}\right) = \frac{-I}{A},\label{eq:ConserveCharge_secondary}
\end{equation}
where $\mathbf{n}_x$ is the normal vector, $I$ is the current demand (in Amps), and $A$ is the battery geometric area.  Equation~\ref{eq:ConserveCharge_secondary} is true
for all locations in the $x$ domain and for all time $t$.  The residual of Eq.~\ref{eq:ConserveCharge_secondary} is added to the PINN loss alongside the typical P2D governing equations residuals (Appendix~\ref{sect:P2DModel_Appendix}).

Second, conservation of charge requires that 
\begin{equation}
    \label{eq:j_int_a}
    \int_{0}^{L_{\rm an}} J_{\rm an} ~{\rm d}x = \frac{-I}{A}.
\end{equation}
and similarly for the cathode, 
\begin{equation}
    \label{eq:j_int_c}
    \int_{L_{\rm an} + L_{\rm sep}}^{L_{\rm an} + L_{\rm sep} + L_{\rm ca}}  J_{\rm ca}~{\rm d}x = \frac{I}{A},
\end{equation}
where $J_{j}$ is the specific production of Li-ions (see Appendix~\ref{sect:P2DModel_Appendix}), and $L_j$ is the thickness of each domain. Enforcing these constraints during PINN training requires spatial integration across each electrode. For every collocation point where the integral constraint is formulated, the spatial integration is achieved with trapezoidal integration using 10 quadrature points uniformly distributed throughout the electrodes. The cost of the regularization residual is therefore typically higher since it requires evaluating the PINN 10 times per collocation point. Therefore, only 640 collocation points are used for the regularization loss, while 2600 collocation points are used for the other losses in the domain interior. Alternative strategies to decrease the cost of the integration are left for future work \cite{saleh2023learning}.

Third, the total amount of Li-ions, at any given time, is constant. Mathematically, this constraint can be expressed as 
\begin{equation}
    \int_{0}^{L_{\rm an} + L_{\rm sep} +L_{\rm ca}} {\epsilon_{{\rm e},j}} c_{\rm e}~ {\rm d}x = C_1,\label{eqn:SaltConserv}
\end{equation}
where $\epsilon_{{\rm e},j}$ is the electrolyte volume fraction and $C_1$ is a constant that can be computed using the initial conditions for $c_{\rm e}$.  Equation~\ref{eqn:SaltConserv} is added to the P2D governing equation set.  

Similar to the weighting procedure for the residuals in Part~I \revone{\cite{hassanaly2023pinn1}}, the additional constraints can be weighted within the PINN. A separate hyperparameter optimization was used to optimize the weights on each additional constraint. Preferentially weighting the conservation given by Eq.~\ref{eq:j_int_a} and Eq.~\ref{eq:j_int_c} significantly improved the accuracy metric $\varepsilon$. The model equipped with secondary conservation constraints is referred to as \textit{Sec.~Cons.}. Figure~\ref{fig:sec_cons}a shows that using the secondary regularization significantly reduces the error of the P2D PINN surrogate.  Figure~\ref{fig:sec_cons}b (bottom) shows examples of the predicted state variables.  Importantly, the added constraints encouraged non-trivial solutions, which allowed the inverting profile to develop for $c_{\rm e}$.
Note that even with secondary conservation, the current $\boldsymbol{i_{\rm e}}$ at $t = 0$ predicted by the PINN can be negative which is non-physical. This can be resolved by using a strict enforcement of initial conditions for the potentials. However, this non-physical behavior only marginally affects the accuracy at later times.

\subsection{Enhanced performance via hierarchical training}

While using secondary conservation constraints improved the PINN accuracy, significant discrepancies between the PINN and a finite-difference solution are still observed (cf., Fig.~\ref{fig:sec_cons}b (bottom)). Given the encouraging multi-fidelity hierarchical training results obtained in Part~I \revone{\cite{hassanaly2023pinn1}}, the same strategy is deployed here. Unlike in Part~I \revone{\cite{hassanaly2023pinn1}}, the P2D surrogate has inaccurate predictions even with a linearized Butler--Volmer formulation. The multi-fidelity hierarchical training proposed here uses the SPM surrogate trained with a linear Butler--Volmer formulation as the first level of the hierarchy (see Part~I \revone{\cite{hassanaly2023pinn1}}). The choice of using the SPM PINN surrogate as the first hierarchy level is motivated by its fast convergence and by the fact that the P2D surrogate model errors are mostly related to its inability to appropriately predict the current density, which is well-captured by the SPM.  Since $c_{\rm e}$ and $\phi_{\rm s,an}$ are not predicted by the SPM model, the hierarchy is only used for $\phi_{\rm s,ca}$, $\phi_{\rm e}$, $c_{\rm s,an}$, and $c_{\rm s,ca}$.

In this hierarchical study, three models are compared: 1) a PINN that uses the proposed hierarchy but no secondary conservation is referred to as \textit{HNN SPM}; 2) a PINN that uses the hierarchy and the secondary conservation referred to as \textit{HNN SPM Sec.~Cons.}; and 3) the \textit{Base} model with secondary conservation and no hierarchy. For the \textit{Base} model, the number of layers and the number of training steps are doubled to provide a fair comparison, which is referred to as \textit{Base Sec.~Cons.~Double}. The cost of training a model with twice as many layers is more expensive than the hierarchy, where one model is frozen while the other is trained. As a result, only three training realizations were done for the \textit{Base Sec.~Cons.~Double} model, which induces large statistical uncertainty on the accuracy metrics. However, since the worst case scenario for \textit{HNN SPM Sec.~Cons.} is more accurate than the best case scenario for \textit{Base Sec.~Cons.~Double}, the statistical uncertainty does not affect the following conclusions. 

\begin{figure}[t!]
    \centering
    \includegraphics[width=2.9in]{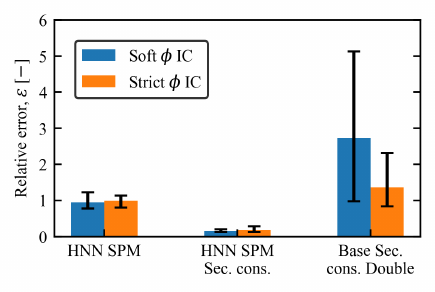}
    \caption{Average PINN error (bar height) for hierarchical training with the SPM as the lower hierarchy level (left), hierarchical training with the SPM as the lower hierarchy level and secondary conservation regularization (middle), and the base model with similar expressiveness as the hierarchical models (right). The error bars denote the 95\% percentile variability observed for all the realizations.}
    \label{fig:ic_hnn_p2d}
\end{figure}

Figure~\ref{fig:ic_hnn_p2d} shows that the best of the \textit{Base} models did not outperform any of the \textit{HNN SPM Sec.~Cons.}~models. Furthermore, using a hierarchy is not a substitute for the secondary conservation regularization as combining the hierarchy and the regularization achieves a higher accuracy than using the hierarchy alone. For all the models, a strict enforcement of the initial conditions for the potential was also used. Figure~\ref{fig:ic_hnn_p2d} suggests that strictly enforcing the initial potentials does not provide a substantial benefit for the accuracy of the PINNs. As a complementary note, the results shown here and in the rest of the paper are obtained with a linearized Butler--Volmer formulation. Using a non-linear formulation instead can simply be done by using an additional hierarchy level such as the one presented in Part~I \revone{\cite{hassanaly2023pinn1}}. The \textit{HNN SPM Sec.~Cons.} models were trained with one additional hierarchy level to account for the non-linear Butler--Volmer formulation. Despite being trained with an additional hierarchy level, the additional error reduction is marginal compared to using secondary conservation constraints (about 12.5\% in the case of the softly enforced initial condition, and 11\% error reduction for the strict enforcement case). 

\begin{figure*}[th!]
    \centering
    \begin{subfigure}[b]{\textwidth}
        \centering
        \includegraphics[width=6.457in]{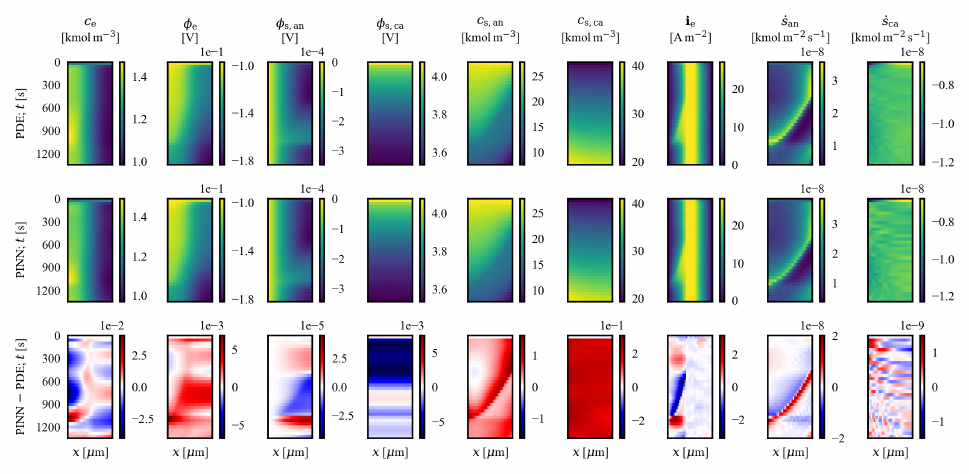}
        \caption{Spatiotemporal contours ($x$-axis is the longitudinal direction, and $y$-axis is the temporal direction from top to bottom) of primary and secondary variables from P2D models. Top: results from PDE integration. Middle: PINN predictions. Bottom: Absolute error between the PDE and PINN solutions. Columns from left to right are for: electrolyte \ce{Li}-ion concentration, electrolyte potential, solid-phase anode potential, solid-phase cathode potential, anode particle surface concentrations, cathode particle surface concentrations, electrolyte-phase current, anode Faradaic current, and cathode Faradaic current.}
        \label{fig:p2d_res_cont}
    \end{subfigure}
    \begin{subfigure}[b]{\textwidth}
        \centering
        \includegraphics[width=6.496in]{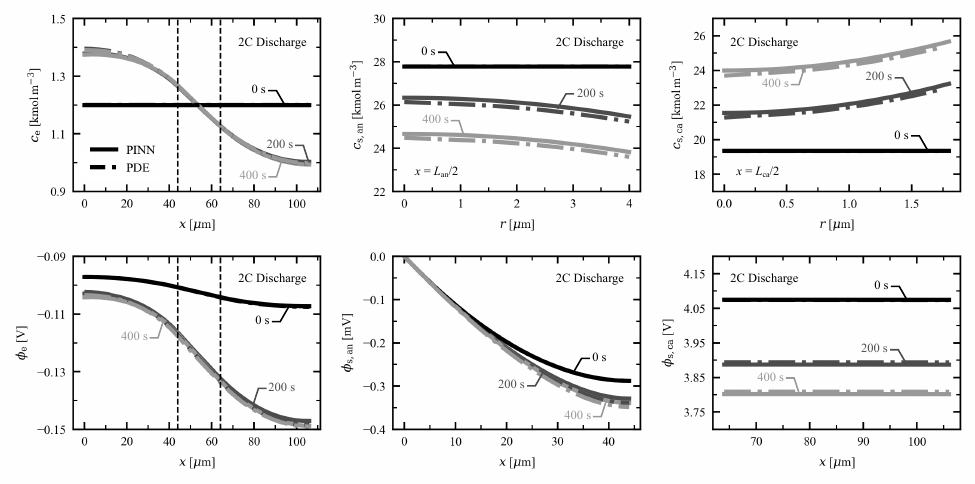}
        \caption{State variable spatial distributions from the PINN and PDE solutions. Profiles are shown for 0~s (\mythickline{black}), 200~s (\mythickline{darkgray}) and 400~s (\mythickline{gray}). Electrode/separator boundaries are shown for the electrolyte \ce{Li}-ion concentration and electrolyte potentials (\mythickdashedline{black}). The variables are the electrolyte \ce{Li}-ion concentration (top left), radial distribution of anode particle \ce{Li} concentrations (top middle), radial distribution of cathode particle \ce{Li} concentrations (top right), electrolyte potential (bottom left), anode potential (bottom middle), and cathode potential (bottom right).}
        \label{fig:p2d_res_line}
    \end{subfigure}

    \caption{Predicted spatiotemporal \ce{Li}-ion battery states using the physics-informed loss from the P2D equations using the \textit{HNN SPM Sec. Cons.} model with strict enforcement potential initial conditions.}
    \label{fig:p2d_res}
\end{figure*}

A P2D PINN surrogate model was selected from the \textit{HNN SPM Sec.~Cons.}~set using strict enforcement of initial potentials. Once selected, training was continued for twice as many epochs. Note that no data loss was used during the training procedure. The results are shown in Fig.~\ref{fig:p2d_res_cont} (middle) and are compared to  results from \textsc{Comsol} PDE integration\footnote{\textsc{Comsol} v6.0; https://www.comsol.com/} (top). The Comsol integration\revtwothree{s} were done with a 30 meshpoints in the radial direction, 20 points throughout the anode and the cathode thicknesses and 5 points throughout the separator. The timestep was dynamically adjusted using a relative tolerance of $10^{-4}$ and absolute tolerance of $10^{-1}$. The PINN surrogate accurately replicates the dynamics and spatial profiles of all dependent variables. Some notable differences can be observed for $c_{\rm e}$, $\phi_{\rm s,an}$ and $\mathbf{i}_{\rm e}$ where the sharp variations at the anode current collector are visibly smoothed by the PINN. Additionally, the temporal variations are slightly slower in the case of the PINN surrogate as compared to the PDE solver, which is evident at early times and results in a temporal shift for the prediction, which is especially apparent for the current density in the anode (Fig.~\ref{fig:p2d_res_cont} bottom). Figure~\ref{fig:p2d_res_line} shows spatial distributions of the state variables. The solid-phase \ce{Li} concentrations $c_{\rm s,an}$ and $c_{\rm s,ca}$ are similar to the results shown in Part~I \revone{\cite{hassanaly2023pinn1}} for the SPM, which justifies using the SPM solution as the first level in the PINN hierarchy. The potentials $\phi_{\rm s,e}$ and $\phi_{\rm s,an}$ quickly deviate and stabilize away from their initial conditions, which may explain why strictly enforcing the initial potential profiles did not have a significant impact on the accuracy. Similar to Part~I \revone{\cite{hassanaly2023pinn1}}, the agreement between the PINN and the PDE solver using only a physics loss validates the implementation and the regularization of the PINN, and its ability to complement the data loss when only sparse data is available.

\section{Parameteric PINN and application to parameter inference}
\label{sec:application}

In this section, the PINN is trained over a multi-dimensional parametric domain, and the benefit of using a PINN for accelerating Bayesian parameter inference is demonstrated. While the computational cost involved in training a PINN to solve the P2D or SPM equations is orders of magnitude higher than the computational cost incurred by using a PDE solver (see Sec.~\ref{sec:timing}), the computational benefit of a PINN best shines when used repeatedly; for instance, when used for parameter inference. In this section, the P2D and SPM surrogates are trained to predict the battery state when using variable battery internal parameters $\mathbf{p}$. In this study, two internal parameters are varied: the exchange current density in the anode $i_{0,{\rm an}}$ and the solid-phase \ce{Li} diffusivity in the cathode $D_{\rm s,ca}$. Note that these parameters depend on state variables (e.g., $c_{\rm e}$ and $c_{\rm s,ca}$), but their variability is modeled with space- and time-independent efficiency parameters $d_{i_{\rm 0,an}}$ and $d_{D_{\rm s,ca}}$, which scale the magnitudes of $i_{\rm 0,an}$ and $D_{\rm s,ca}$, respectively. The efficiency parameters are calibrated and are provided as the input of the PINN surrogate. The choice of these internal parameters is motivated by aging mechanisms linked to solid-electrolyte interface (SEI) degradation which affects $i_{\rm 0,an}$, and cathode cracking that affects $D_{\rm s,ca}$. Both internal parameters are modeled by the SPM and the P2D equations, which allows using the same analysis with both surrogates. In the following, the prior distribution $p_{\rm prior}$ of $d_{D_{\rm s,ca}}$ is a uniform distribution $\mathcal{U}(1, 10)$ and the prior distribution of $d_{i_{\rm 0,an}}$ is $\mathcal{U}(0.5, 4)$. The ranges are unitless since the efficiency parameters scale the original parameters.

\subsection{Internal property variation}
\label{sec:parametericPinn}

Modeling the effect of $d_{i_{\rm 0,an}}$ and $d_{D_{\rm s,ca}}$ on the PINN solution requires including the efficiency parameters $d_{i_{\rm 0,an}}$ and $d_{D_{\rm s,ca}}$ as input variables of the PINN. Since the input domain dimension is expanded, additional collocation points are needed to fully cover the spatiotemporal and parametric space. Assuming a tensor expansion for the number of collocation points that span the domain, one can quickly see how relying solely on collocation points will expose the PINN to the curse of dimensionality as the number of internal parameters included increases. Two methods to mitigate this issue are explored here. First, a hierarchical approach is used where the solution at one parameter set (here $(d_{i_{\rm 0,an}}, d_{D_{\rm s,ca}}) = (0.5, 1.0)$) is used as the first level of the hierarchy. The second level of the hierarchy learns to correct that solution to adapt to different parameter sets. This approach uses the fact that the sharp temporal and spatial variations occur at similar spatial and temporal locations for all calibrated parameter values. Note that using the spatiotemporal features of one solution to inform the spatiotemporal feature of another is part of the recent success of the operator learning method, where an explicit separation between spatiotemporal and parametric training is enforced~\cite{pan2022neural,de2023bi,lu2021learning}. For the P2D model, the multi-fidelity hierarchical approach results in the interaction of 4 distinct neural networks: 
\begin{itemize}
    \item One for capturing the initial conditions at the parameter set $(0.5, 1.0)$; \item One for the SPM solution at the parameter set $(0.5, 1.0)$;
    \item One that uses the two previous neural nets to predict the P2D solution at the parameter set $(0.5, 1.0)$; and
    \item One that learns the solution of the P2D equations for all parameter values. 
\end{itemize}

\begin{figure}[t!]
    \centering
    \includegraphics[width=2.403in]{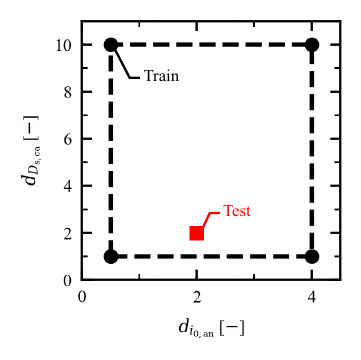}
    \caption{Full parametric domain space (\mythickdashedline{black}), parameters included in the training data (\mythickcircle{black}{black}), and parameter set reserved for testing (\mythicksquare{red}{red}).}
    \label{fig:parametricDomain}
\end{figure}

Given that strict enforcement of initial conditions did not significantly improve PINN accuracy, the initial conditions are only softly enforced. Second, instead of solely relying on physics loss, one can also use data from the PDE solver obtained for some parameter set in the domain. In the case of the parameter calibration task (instead of the PINN regularization task), using data will also allow evaluation of whether data can altogether be replaced by physics loss. The number of data sets used here is limited to four, which corresponds to two observations per calibrated parameter. For the 2D parameter domain considered here, one could theoretically generate a larger amount of PDE solutions that would fully cover the calibrated parameter space, and not require physics loss. However, this approach is not feasible as the dimension of the parameter space increases. Assuming a $22$-dimensional parameter space, which is half of the internal parameters considered in Reddy et al.~\cite{reddy2019accelerating}, a 360~s wall time to compute a single PDE solution, three PDE solutions per dimension would require about $10^{9}$ CPUh and petabytes of storage. Limiting the analysis to two PDE solutions per dimension aims to mimic the sparse availability of data that can be expected.
Ideally, data should be available at the values $\{(0.5, 1.0), (0.5, 10.0), (4.0, 1.0), (4.0, 10.0)\}$, i.e., at the corners of the parameter space. A PDE solution at $(2.0, 2.0)$ is also generated to serve as test data to test the ability of the PINN to capture battery behavior in unseen conditions.  Figure~\ref{fig:parametricDomain} illustrates the parametric domain. 

For both the SPM and the P2D model, six PINN surrogate models are compared as an ablation study: 
\begin{itemize}
    \item[1)] A model using a hierarchy, physics loss, and data loss denoted as $HNN + phys. + data$;
    \item[2)] A model using the hierarchy and the physics loss only denoted as $HNN + phys$;
    \item[3)] A model using the hierarchy and only the data loss denoted as $HNN + data$;
    \item[4)] A model using only the physics loss and no hierarchy nor data loss denoted as $phys$;
    \item[5)] A model using the physics loss and the data loss but no hierarchy denoted as $phys. + data$; and
    \item[6)] A model that uses only the data loss denoted as $data$.
\end{itemize}
Note that in the case of the P2D model, ``physics loss'' contains the secondary conservation regularization. Both SPM and P2D models trained for parameter calibration use a linear Butler--Volmer formulation as a proof-of-concept. A hierarchical approach to handling the non-linear Butler--Volmer formulation is outlined in Part~I \revone{\cite{hassanaly2023pinn1}}. The models are evaluated against the PDE solution by computing an error $\varepsilon$ with Eq.~9 in Part~I \revone{\cite{hassanaly2023pinn1}} for the SPM and Eq.~\ref{eq:errP2D} for the P2D model.

\begin{figure*}[th!]
    \centering

    \includegraphics[width=6.376in]{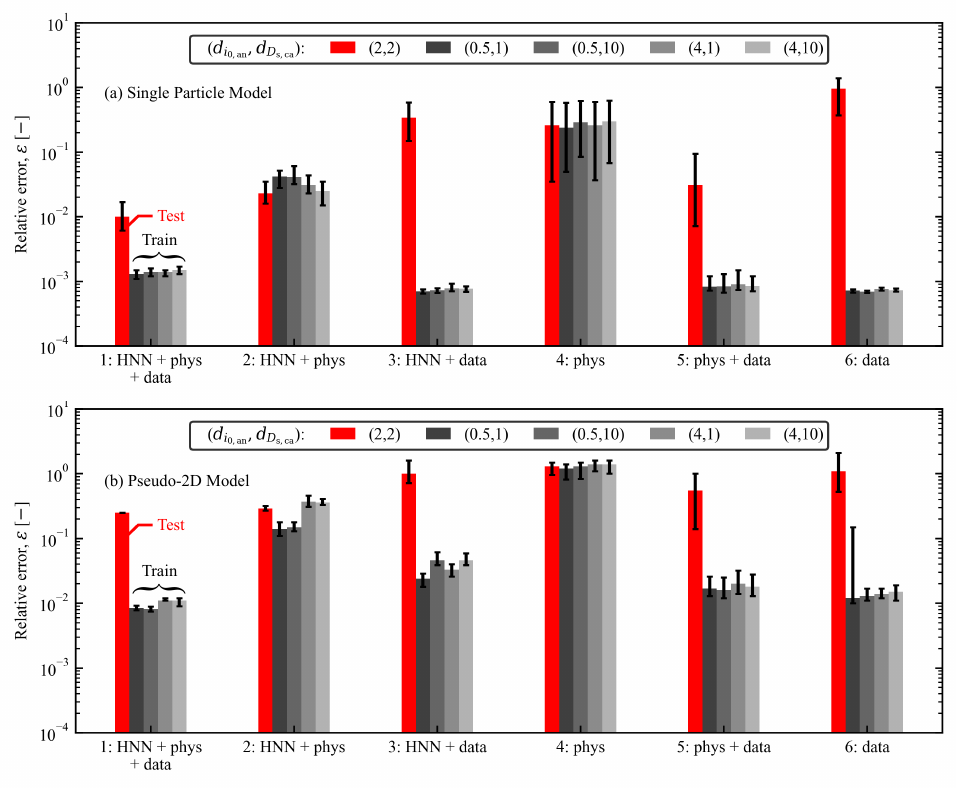}
    \caption{Average PINN error (bar height) over the parameter range considered for the exchange current density in the anode $i_{\rm 0,an}$ and the \ce{Li} diffusivity in the cathode $D_{\rm s,ca}$. The red bar shows the accuracy result for a parameter set not included in the data set. The black bars show the accuracy for the parameter sets included in the data set. The error bar denotes the 95\% percentile variability observed for all the realizations.}
    \label{fig:par_var}
\end{figure*} 

Figure~\ref{fig:par_var}a (resp.~Fig.~\ref{fig:par_var}b) shows the training results for the SPM (resp.~P2D). The $data$ models show for both the SPM and P2D cases that high accuracy is achieved where data is available. However, the error can be 2--3 orders of magnitude higher in regions where no data is available. These observations justify the procedure undertaken here (in training using physics loss) as one cannot solely rely on the interpolation of a neural net between locations where data is available. Comparing the $phys. + data$ and $data$ models indicates that the error incurred where no data is available decreases when physics loss is enforced, while lower accuracy is achieved where data is available. The role of the physics loss is to ensure that the interpolation between the available data does not violate the governing equations, which results in a higher accuracy where no data is available. Comparing the $phys.$ and $phys. + data$ models highlights the importance of data. While the physics loss alone can replicate the PDE solution, it is more efficient to use data if it is available. 
This observation highlights that although the physics loss can replace data, the physics loss is best used as a complement to data rather than as a complete replacement. Comparing the $HNN + data$ and $data$ models, the hierarchy is only slightly helpful in the absence of a physics loss to predict the solution where data is not available. When data is available, using the hierarchy increases the error incurred. This can be explained by the fact that although the hierarchy helps steer the physics loss away from trivial solutions, it may also reduce the expressiveness of the neural network. This is the manifestation of the intended behavior of the hierarchical approach where the higher hierarchy levels seek small corrections to the solutions at the lower hierarchy levels. Comparing the $HNN + phys.$ and $phys.$ models, it is clear that the hierarchical approach allows for a higher effectiveness of the physics loss, in general. Comparing the $HNN + phys.$ and $phys. + data$ models highlights that although the hierarchical approach helps with the accuracy obtained with the physics loss, it is not a substitute for using less data. In other words, if data is available, it should be used. This conclusion is strengthened when comparing the $HNN + phys. + data$ and $HNN + phys.$ models, where higher accuracy is achieved overall if data supplements the hierarchy and the physics loss. \revtwotwo{The errors discussed in this section evaluate the accuracy of all the state variables. A similar analysis is provided in Appendix~\ref{sect:termVoltErr} for the errors in the terminal voltage, which is a typical quantity used for model calibration Sec.~\ref{sec:cal}.}

\subsection{Parameter calibration}
\label{sec:cal}

Given an accurate surrogate model that captures the effect of the variability in several internal parameters, the surrogate can be used to infer internal parameters from partial experimental observations (i.e., the battery voltage response). As mentioned in Section~\ref{sect:Introduction}, multiple sources of uncertainty can affect the parameter inference, which motivates the use of Bayesian calibration. Here, the \verb|numpyro| implementation of the no U-turn sampling variant of Hamiltonian Monte-Carlo (HMC) is used~\cite{hoffman2014no, numpyro}. In short, HMC is a sampling method that uses the gradient of the likelihood function with respect to the parameters being inferred to guide sampling. As a result, it requires minimal chain warm-up and allows for a high sample acceptance probability (set to 0.9 here). This method is particularly attractive in the present context given that gradients of the solution with respect to the parameters calibrated can be readily computed with the neural net surrogate, which is, in general, not possible for PDE solvers. 

The variable $\phi_{\rm s, ca}$ (i.e., the battery voltage) is used as the observation for the unseen parameter set $(2.0, 2.0)$ over the time interval $[0~\rm{s}, 1350~\rm{s}]$ during a 2~C discharge. The PDE integration is replaced by a PINN surrogate that uses physics loss and a data loss (Model~1) to compute the likelihood function $p_{\rm like}$. The standard deviation $\sigma$ in the likelihood function (Eq.~\ref{eq:like}) is chosen to be uniform across the observations and is set by a hyperparameter search within the bounds $[1~{\rm mV}, 100~{\rm mV}]$. The chosen uncertainty value is the minimal value that ensures that 95\% of the predictions obtained with the surrogate model evaluated at the sampled parameter values are contained within two standard deviations of the observation data. This approach lumps the observation uncertainty and the model uncertainty (inaccuracy of the PINN surrogate) within the $\sigma$ value of the likelihood function (Eq.~\ref{eq:like}). The first 10,000 samples are discarded as part of the Markov chain warm-up and the next 4,000 are selected as the posterior samples. Two calibration scenarios are considered: in the first scenario, the observation data is the unaltered result of the PDE integration. In the second scenario, the observation data is superimposed with a noise normally distributed with a standard deviation of $3~\rm{mV}$. The purpose of the noisy scenario is to highlight the benefit of using a Bayesian calibration procedure in the presence of experimental uncertainty.

\begin{figure*}
    \centering
    \includegraphics[width=6.604in]{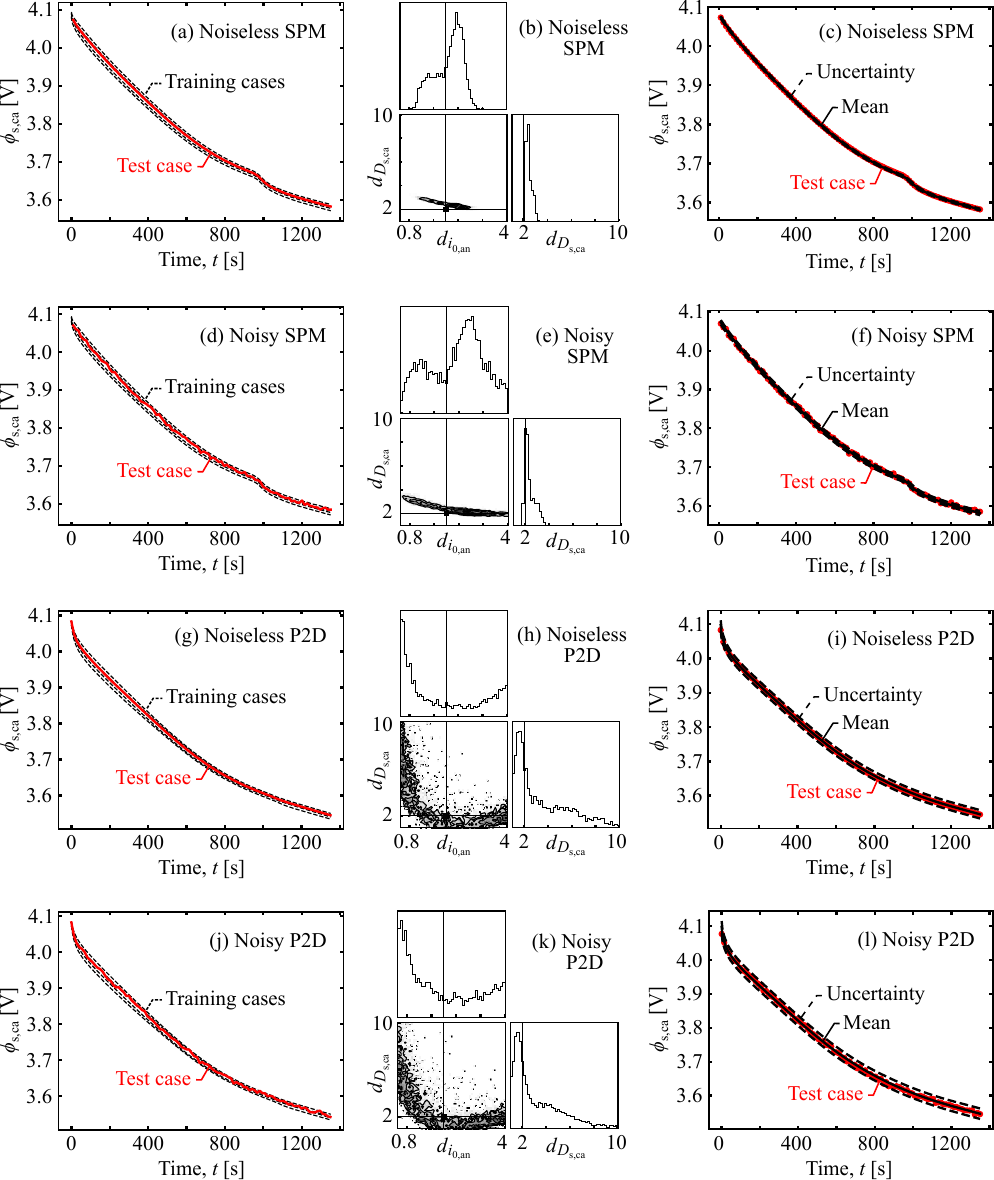}
    \caption{Parameter calibration results for the PINN surrogate for the SPM model (a-f) and the P2D model (g-1).  Shown are the calibration results for the noiseless cases (a-c, g-i) and the noisy cases (d-f, j-l). The left figures show the cathode potential at the current collector (i.e., the battery voltage) for the trained cases and the test case.  The middle figures show a corner plot of the calibrated parameters. The right figures show the predicted cathode potential at the current collector $\phi_{\rm s,cc}$ obtained with the mean posterior samples along with responses from parameters at the 95\% percentile bounds.}
    \label{fig:spm_p2d_cal}
\end{figure*}

The Bayesian calibration results using the SPM surrogate are shown in Fig.~\ref{fig:spm_p2d_cal}a-f. For the noiseless scenario (Fig.~\ref{fig:spm_p2d_cal}a-c), the best likelihood standard deviation $\sigma$ was found to be $2.0~\rm{mV}$. The variability in the observation data is illustrated by plotting the positive electrode profile at the edge of the parameter space. While varying the internal battery parameters does have an effect on the observation data, the effect is contained within about $20~\rm{mV}$, which requires a high accuracy from the surrogate model. The calibration procedure successfully identifies that parameters near the value $(2.0, 2.0)$ explain the observation data. While the model is highly confident in the value of the $d_{D_{\rm s,ca}}$, the uncertainty in $d_{i_{\rm 0,an}}$ is about twice as large, which suggests that identifying the exchange current density accurately would require a more accurate surrogate or different observation data~\cite{VSSAYM17}. The difference in identifiability between $d_{i_{\rm 0, an|}}$ and $d_{D_{\rm s, ca}}$ points to a high level of parameter sloppiness~\cite{dufresne2016geometry,MLDBBBVMMA22} (i.e., the parameters inferred are not equally sensitive to the voltage response). Finally, the forward surrogate model evaluation at the posterior samples closely predicts the observed battery voltage (right).

In the case of noisy observations (Fig.~\ref{fig:spm_p2d_cal}d-f), the best likelihood uncertainty is found to be $5.36~\rm{mV}$, which is consistent with the size of the noise introduced in the observations. In turn, the uncertainty in the predicted parameters increases, especially for the exchange current density, which is subject to high uncertainty. Again, despite the high uncertainty values, the forward model evaluation closely matches the observation data. The ability of the Bayesian calibration procedure to reflect uncertainties into the inferred parameter values is the main reason behind using the present method in place of a non-Bayesian optimization procedure. Further validation of the support of the posterior distribution is provided in Appendix~\ref{sect:SPMBrute}. 

\begin{table*}[t]
\scriptsize
\setcellgapes{5pt}
\makegapedcells
\caption{Computational cost breakdown of the entire calibration procedure for the P2D model and the SPM. The computational cost is expressed in total computing time and ``Time Equivalent PDE'' which gives the number of forward evaluations that could have been done during the calibration.}
\vspace*{-5mm}
\label{tab:timing}
\begin{center}
\begin{tabular}{ |c|c|c|c|c|c|c|c|c| } 
\hline
& & \multicolumn{4}{|c|}{\textbf{Training}} & & \multicolumn{2}{|c|}{\textbf{Total}} \\
\hline
& Data Gen. & SPM (0.5, 1) & IC (0.5, 1) & P2D (0.5, 1) & Full param. set & MCMC & Time & Time Eq. PDE \\
\hline
P2D & $1,440~{\rm s}$ & $3,674~{\rm s}$ & $310~{\rm s}$ & $17,274~{\rm s}$ & $44,985~{\rm s}$ & $51~{\rm s}$ & $67\times 10^3~{\rm s}$ &  $186$ \\
\hline
SPM & $120~{\rm s}$ & $3,674~{\rm s}$ & - & - & $12,338~{\rm s}$ & $44~{\rm s}$ & $16\times 10^3~{\rm s}$ &  $530$ \\
\hline
\end{tabular}
\end{center}
\end{table*}

Figure~\ref{fig:spm_p2d_cal}g-l shows the same experiments conducted for the P2D model. Just like in the SPM case, the variability of $\phi_{s,cc}$ with respect to changes in the parameters inferred is small. In the P2D case, the errors were typically larger than for the SPM case throughout the parameter set. As a result, even in the noiseless case (Fig.~\ref{fig:spm_p2d_cal}g-i), the best likelihood uncertainty value was found to be $10.2~\rm{mV}$, which is higher than for the SPM case and illustrates the higher model errors of the P2D surrogate. This observation suggests that compared to the PINN SPM surrogate additional PDE data should be used to train the PINN P2D surrogate. The joint PDF of the posterior sampled in the parameter space has a structure similar to the SPM case but has a larger support, which is a result of the higher uncertainties. Nevertheless, the value of $d_{D_{\rm s,ca}}$ is confidently identified. In the noisy case (Fig.~\ref{fig:spm_p2d_cal}j-l), the best likelihood uncertainty is found to be $10.4~\rm{mV}$, which is only mildly larger than in the noiseless case. The similar uncertainty value predicted suggests that the main source of uncertainty\revtwothree{,} in this case\revtwothree{,} is the model uncertainty rather than the observational uncertainty, and adding noise to the data does not hurt or improve the parameter inference. As a result, the posterior PDF of the parameter inferred closely resembles the PDF obtained in the noiseless case.

\subsection{Computational cost}
\label{sec:timing}

Although using surrogate models might incur higher uncertainty during parameter inference, as shown in Sec.~\ref{sec:cal}, the main advantage of the procedure is its lower computational cost. In this section, computational savings are described. An ``apples-to-apples'' computational cost comparison is difficult for several reasons described hereafter. The PINN is being trained for a fixed number of epochs to compare the performances across the models used. Early stopping could be used to reduce the computational cost, while inducing minimal effects on accuracy. Likewise, the PINN architecture and the number of collocation points could be reduced to decrease the training procedure computational cost. On the PDE solver side, several acceleration methods through simplification of the governing equations could be used to speed-up the PDE evaluations, while also incurring minimal errors. The number of samples drawn with the MCMC procedure could be minimized to reduce the number of PDE solves. 

\begin{table}[t]
\scriptsize
\setcellgapes{5pt}
\makegapedcells
\caption{Computational cost to run MCMC using the PINN surrogates and PDE solutions.}
\vspace*{-5mm}
\label{tab:timing2}
\begin{center}
\begin{tabular}{ |l|c|c| } 
\hline
 Model & \shortstack{Forward simulations\\ for MCMC} & \shortstack{MCMC \\computational time} \\
\hline
SPM PDE & 30~s & 12.6~$\times 10^6$s \\
\hline
PINN SPM surrogate & 314~$\mu$s & 44~s \\
\hline
P2D PDE & 360~s & 151.2~$\times 10^6$s \\
\hline
PINN P2D surrogate & 364~$\mu$s & 51~s \\
\hline
\end{tabular}
\end{center}
\end{table}

For the SPM PDE solver, a custom implicit solver is used (available in the companion repository) and requires about $30$~s per parameter evaluation \revone{when using $N_{\rm r} = 32$ and $\Delta_t = 0.75$~s, where $N_{\rm r}$ is the number of mesh points throughout the electrode particles and $\Delta_t$ is the timestep.} For the P2D solver, the \textsc{Comsol} PDE solver took on average $360$~s per parameter evaluation. During the MCMC procedure, the calibration is repeated at most 10 times to optimize the uncertainty used in the likelihood. The computational cost of the PINN training and calibration procedures are shown in Table~\ref{tab:timing}\footnote{The runs were done Dual Intel Xeon Gold Skylake 6154 3.0 GHz processors}. For both the P2D and the SPM models, the vast majority of the computing time is spent in training the model that encodes the effect of parameter variations. In particular, the data generation accounts for at most 2\% of the total computing time. The total time spent is shown in terms of an equivalent number of PDE solves. For the entire calibration procedure against measurement of a single discharge cycle, $140,000$ MCMC samples were drawn. For each PINN evaluation, the gradient of the voltage response with respect to the inferred parameters, needed by the HMC procedure, can be obtained via auto-differentiation \cite{baydin2018automatic}. For a PDE solver, this gradient can be obtained with $N+1$ PDE solves, where $N$ is the number of parameters inferred, here $N=2$. This results in a speed-up of $2250$x for the P2D surrogate and $780$x for the SPM surrogate (Table~\ref{tab:timing2}).

The calibration step accounts for at most 0.3\% of the total computational time, while it is the main cost for calibrating with a PDE solver. Therefore, once a PINN surrogate is trained, it can be reused for other cycle numbers and other cells with minimal overhead. The actual speed-up for degradation modeling is likely to be orders of magnitude higher.

Finally, in terms of memory requirements, the SPM surrogate including the entire hierarchy contains $18,008$ parameters, while the P2D surrogate with the entire training hierarchy contains $38,458$ parameters. Recasting the parameter into single precision after training requires $72$~kB and $153$~kB, respectively. Given the low computational intensity of the MCMC procedure, the parameter inference could possibly be done onboard electrified devices. 

\revall{
Table~\ref{tab:arch_spm} summarizes the SPM surrogate neural net architecture. Similar to Part~I~\cite{hassanaly2023pinn1}, the merged part of the network contains a single fully connected layer (denoted by FC in Tab.~\ref{tab:arch_spm}) in the block that connects to the parameters, the time variable $t$, and the radial spatial variable $r$. The branches (that connect the merged part of the network to the predicted state variables), use three gradient pathology blocks (denoted by GP in Tab.~\ref{tab:arch_spm}) with 20 neurons per layer. The first and second level\revtwothree{s} of the hierarchy use the same architecture where each contains $9,004$ trainable parameters. Therefore, the entire SPM surrogate contains $18,008$ trainable parameters.}

\begin{table}[!hbt]
    \centering
    \scriptsize
    \caption{\revall{SPM surrogate architecture.}}
    \revall{\begin{tabular}{c|l|l}
          & SPM (0.5, 1) & SPM Full par. \\
         \hline
         Neuron per layer & 20 & 20  \\
         Activation & $\operatorname{tanh}$ & $\operatorname{tanh}$ \\
         Merged part & 1 (FC) & 1 (FC) \\
         Branch part & 3 (GP) & 3 (GP) \\      
    \end{tabular}}
    \label{tab:arch_spm}
\end{table}

\revall{
Table~\ref{tab:arch_p2d} summarizes the  P2D surrogate neural net architecture. The surrogate uses the first level of the SPM surrogate as its first hierarchy level. The initial conditions of the potentials are learned with a small network that uses a single fully connected layer for the merged part of the network and a single fully connected layer for the branch that connects the merged part to the potentials. In total, the initial condition network uses $1,242$ trainable parameters. The P2D solution at parameters (0.5, 1) and the network that encodes the full parameter space use two fully connected layers as the merged part of the network and three gradient pathology blocks in the branching part of the network. The activation\revtwothree{s} of the final layers are the same as in the SPM surrogates for $c_{\rm s, an}$,  $c_{\rm s, ca}$, $\phi_{\rm e}$ and $\phi_{s, ca}$. Linear activations are additionally used for $c_{\rm e}$ and $\phi_{s, an}$. Both networks use $14,106$ trainable parameters. In total, the P2D surrogate uses $38,458$ trainable parameters.} 

\revall{
\begin{table}[!hbt]
    \centering
    \scriptsize
    \caption{\revall{P2D surrogate architecture.}}
    \revall{\begin{tabular}{c|l|l}
          & IC (0.5, 1) &   P2D (0.5, 1)/ Full par. \\
         \hline
         Neuron per layer & 20 & 20  \\
         Activation & $\operatorname{tanh}$ & $\operatorname{tanh}$ \\
         Merged part & 1 (FC) & 2 (FC) \\
         Branch part & 1 (FC) & 3 (GP) \\      
    \end{tabular}}
    \label{tab:arch_p2d}
\end{table}}

\section{Discussion}
\label{sect:Discussion}
This work is a continuation of the Part~I paper \revone{\cite{hassanaly2023pinn1}} and highlights key similarities and differences between training a PINN for an SPM or a P2D model. As in the SPM model, the PINN can fail by only finding the trivial (no change from initial conditions) solution. In the case of the SPM, this issue was avoided by weighting the particle surface boundary condition. In the case of the P2D equations, weighting the residuals alone can fail because the current density is not held fixed. In this work, adding secondary conservation constraints successfully avoids this training failure mode and encourages the PINN to be trained with only physics-informed losses.

The secondary conservation is not a substitute for the multi-fidelity hierarchical training introduced in Part~I \revone{\cite{hassanaly2023pinn1}}, which is shown to perform best when used in tandem with the secondary conservation regularization loss. A remarkable conclusion is that for the P2D model, the lower hierarchy levels can be formulated with the solution of the SPM. First, the success of this approach suggests that using other modeling fidelities of Li-ion batteries could be a viable option for further improvement~\cite{moura2016battery}. Second, using the SPM solution as a base for the P2D solution effectively implemented a tensorial expansion from two spatio-temporal variables ($t,r$) to three ($t,r,x$). Potentially, the multi-fidelity training could apply to tensorial expansions in the parametric domain and allow the PINN to be more efficiently trained in higher dimensions. 

The practical benefits of the trained surrogates is demonstrated for the first time in the context of \ce{Li}-ion battery parameter inference. The results presented in Section~\ref{sec:application} suggest that solely relying on interpolating a neural network between available data can lead to inaccurate results where data is not available. Using the physics-informed loss helps achieve higher accuracy where data is not available. Nevertheless, using data and the physics loss is the best combination to achieve high accuracy throughout the parameter space. Designing a PINN regularization for battery models should therefore not supplant data generation. Using a hierarchical training formulated via tensorial expansion of the parameter space was beneficial, especially when used with a physics and data loss. The regularization method for the PINN training does not act as a substitute for training without data, but rather complements data to make the Bayesian parameter inference possible in high-dimensions.

For parameter inference, the PINN surrogate provides a fast calculation of the likelihood function (Eq.~\ref{eq:like}) and enables calculating gradients with respect to the inferred parameter values, thereby enabling efficient MCMC methods. Past the initial investment of the PINN training, the surrogate decreases the cost of model predictions by several orders of magnitude, making MCMC a cheap process that can be used for offline and online diagnostics.

\revtwo{In the framework proposed, parameter inference in high dimensions (10 to 40) would expose the method to the curse of dimensionality, as the number of collocation points needed to span the entire parameter space exponentially increases with parameter dimension. This is especially problematic given that a full batch L-BFGS training is used in the second part of the training process. Scaling scientific machine learning methods to enable uncertainty quantification in high dimensions is a known challenge that is receiving increasing attention~\cite{psaros2023uncertainty}. An immediate solution would consist in reducing the size of the parameter space spanned by the collocation points. Dimension reduction of the parameter space can be envisioned either via expert knowledge or by calibrating groups of parameters~\cite{namor2017parameter,jobman2015identification}. Another way to combat the curse of dimensionality is to leverage the hierarchical training method outlined in Part~I (i.e., instead of training a PINN over the full parameter space, one could envision combining multiple PINNs trained over a subset of the parameters). Those strategies will be explored in future work.}

\revtwo{Finally, the trained PINNs were, so far, demonstrated with constant-current discharge only. While there is no inherent limit to applying the same method for charging \revtwotwo{(see Appendix~\ref{sect:P2DModel_Appendix} in Part I)}, there could be complications if the current is not known a priori. In particular for constant-voltage and constant-power conditions, the regularization of Eq.~\ref{eq:j_int_a} and Eq.~\ref{eq:j_int_c} are not applicable since their right-hand side is not known. In those cases, a heavier reliance on data or hierarchical training may be required. Time-dependent duty cycles are also not an inherent limitation \revtwotwo{(see Appendix~\ref{sect:P2DModel_Appendix} in Part I)}. However, if the PINN needs to adapt to multiple duty cycles, then the input space might, again, be large enough to expose the PINN to the curse of dimensionality. In such cases, a physics-informed neural operator approach might be more appropriate to span the infinite-dimensional space of time-dependent duty cycles~\cite{goswami2023physics}. These issues will also be explored as part of future investigations. }

\section{Conclusion}
\label{sect:Conclusion}

In this work, an approximation of a \ce{Li}-ion battery P2D model was obtained for the first time with a PINN. Using secondary conservation regularization and a hierarchical training method derived in Part~I \revone{\cite{hassanaly2023pinn1}}, it was shown that the P2D model solution could be well-approximated without data. The data-efficiency is a key feature of the PINN, which can enable building a surrogate for high-dimension problems with sparse data availability. The PINN was adapted to predict the dependence of the state of \ce{Li}-ion batteries with respect to unknown internal parameters. Combining physics and data losses leads to the highest accuracy throughout the explored parameter space. The surrogate model was successful at accelerating Bayesian parameter inference, while leading to a prediction that was two orders of magnitude faster than a PDE solver, including training time.  Future work will be dedicated to scaling up the approach to higher-dimension parametric spaces either via multi-fidelity training and/or operator learning. 

\section*{Acknowledgements}

This work was authored by the National Renewable Energy Laboratory (NREL), operated by Alliance for Sustainable Energy, LLC, for the U.S. Department of Energy (DOE) under Contract No. DE-AC36-08GO28308.  This work by authored in part by Idaho National Laboratory (INL) operated by Battelle Energy Alliance, LLC under contract No. DE-AC07-05ID14517. This work was supported by funding from DOE's Vehicle Technologies Office (VTO) with Simon Thompson as program manager and DOE's Advanced Scientific Computing Research (ASCR) program with Steven Lee as program manager. A.D.'s work was also partially supported by the AFOSR awards FA9550-20-1-0138. The research was performed using computational resources sponsored by the Department of Energy's Office of Energy Efficiency and Renewable Energy and located at the National Renewable Energy Laboratory. The views expressed in the article do not necessarily represent the views of the DOE or the U.S. Government. The U.S. Government retains, and the publisher, by accepting the article for publication, acknowledges that the U.S. Government retains a nonexclusive, paid-up, irrevocable, worldwide license to publish or reproduce the published form of this work, or allow others to do so, for U.S. Government purposes.

\section*{Nomenclature}
\scriptsize
\begin{tabbing}
	Variable \hspace{5mm}\= Description \hspace{35mm} \= SI Units \\
        $A$                     \> Battery geometric area \> m$^2$\\
        $c_{\rm e}$             \> Electrolyte Li-ion concentration \> kmol~m$^{-3}$\\
        $c_{{\rm s},j}$         \> Solid-phase Li concentration in phase $j$\>  kmol~m$^{-3}$\\
        $c_{{\rm s, max},j}$    \> Max solid-phase Li concentration in phase $j$\> kmol~m$^{-3}$\\
        $C_1$                   \> Initial salt concentration \> kmol~m$^2$\\
        $\boldsymbol{d}$            \> Observations \>  \\
        $\boldsymbol{d}_{\rm pred}$            \> Predicted observations \>  \\
        $d_{i_{\rm 0,an}}$      \> Anode exchange current density efficiency factor \> $-$\\
        $d_{D_{\rm s,ca}}$      \> Ca active material diff. coeff. efficiency factor \> $-$\\
        $D_{{\rm e}}$           \> Bulk-phase Li-ion diffusion coefficient \> m$^2$~s$^{-1}$\\
        $D^{\rm eff}_{{\rm e},j}$   \> Effective Li-ion diffusion coefficient \> m$^2$~s$^{-1}$\\
        $D_{{\rm s},j}$         \> Solid-phase Li diffusion coefficient \> m$^2$~s$^{-1}$\\
        $D_{\rm t}$             \> Ramping function in time \> $-$ \\
        $D_{\rm x}$             \> Ramping function in space \> $-$\\
        $F$                     \> Faraday's constant \> s~A~kmol$^{-1}$\\
        $i_0$                   \> Exchange current density \> A~m$^{-2}$ \\
        $i^0_0$                 \> Exchange current density prefactor \> A~m$^{-2}$ \\
        $\mathbf{i}_{\rm e}$    \> Electrolyte current density \> A~m$^{-2}$ \\
        $I$                     \> Current demand \> A~m$^{-2}$\\
        $j$                     \> Phase indicator \> $-$ \\
        $J_j$                   \> Li-ion flux due to electrochemical reactions \> kmol~m$^{-2}$~s$^{-1}$\\
        $L_{\rm an}$            \> Anode thickness \> m \\
        $L_{\rm ca}$            \> Cathode thickness \> m \\
        $L_{\rm sep}$            \> Separator thickness \> m \\
        $N_{\rm d}$             \> Number of observations \> $-$\\
        $\boldsymbol{p}$            \> Parameter set \>  \\
        $p_{\rm like}$          \> Likelihood function \> $-$ \\
        $p_{\rm post}$          \> Posterior probability function \> $-$ \\
	$p_{\rm prior}$         \> Prior probability function \> $-$ \\
        $p_{{\rm s},j}$         \> Solid-phase Bruggeman coefficient \> $-$\\
        $p_{{\rm e},j}$         \> Liquid-phase Bruggeman coefficient \> $-$\\
        $r$                     \> Radial coordinate \> m \\
        $R$                     \> Universal gas constant \> J~kmol$^{-1}$~K$^{-1}$\\
        $R_{{\rm s},j}$         \> Representative active material particle radius \> m\\
        $\dot{s}_j$             \> Production rate of Li-ions \> kmol~m$^{-2}$~s$^{-1}$\\
        $t$                     \> Time \> s \\
        $t^0_{+}$               \> Li-ion transferrence number \> $-$ \\
        $T$                     \> Temperature \> K \\
        $U_{{\rm OCP},j}$       \> Open-circuit potential of active material $j$ \> V\\
        $x$                     \> Through-plane coordinate \> m\\
        $\alpha_{\rm a}$        \> Anodic symmetry factor \> $-$ \\
        $\alpha_{j}$            \> Phase concentration scaling factor \> kmol~m$^{-3}$\\
        $\epsilon_{{\rm AM}, j}$ \> Active material volume fraction in phase $j$ \> $-$\\ 
        $\epsilon_{{\rm e},j}$  \> Electrolyte volume fraction in phase $j$ \> $-$ \\
        $\epsilon_{{\rm s}, j}$ \> Solid-phase volume fraction in phase $j$ \> $-$\\ 
        $\varepsilon$           \> Scaled mean aboslute error \> $-$ \\ 
        $\eta_j$                \> Kinetic overpotential \> V\\
        $\xi_{\rm m}$           \> Predicted state variable from model $m$\> \\
        $\widetilde{\xi}$       \> Raw predicted state variable from neural net\> \\
        $\sigma$                \> Standard deviation \> \\
        $\sigma_{{\rm s},j}$  \> Bulk-phase solid-phase conductivity \> S~m$^{-1}$\\
        $\sigma^{\rm eff}_{{\rm s},j}$  \> Effective solid-phase conductivity \> S~m$^{-1}$\\
        $\tau$                  \> Timescale with significant init.~condition effect \> s\\
        $\phi_{\rm e}$          \> Electrolyte potential \> V\\
        $\phi_{{\rm s},j}$      \> Solid-phase potential in composite $j$\> V \\
\end{tabbing}

\bibliographystyle{unsrt}

\section*{Appendix}
\appendix
\renewcommand{\thesection}{\Alph{section}}
\setcounter{section}{0}
\setcounter{figure}{0}
%
\small

\section{P2D model governing equations}
\label{sect:P2DModel_Appendix}
\input{P2DGovEqns}

\section{Terminal voltage errors}
\label{sect:termVoltErr}
\input{part2_termvolt}

\section{PDE-based calibration}
\label{sect:SPMBrute}
\input{SPM_Brute_revised_highlights}

\end{document}

%% file: P2DGovEqns.tex
In the pseudo-2D (P2D) model, the electrolyte is represented as a dual-salt concentrated solution~\cite{FDN94}.  Unlike the single particle model (SPM), the P2D model does not assume that the electrolyte is ideal and instead resolves the Li-ion and potential gradients.  Li-ion conservation within the electrolyte can be expressed as 
\begin{equation}
	\frac{\partial \left(\epsilon_{{\rm e},j}~ c_{\rm e}\right)}{\partial t} = \nabla_x\cdot \left(D^{\rm eff}_{{\rm e},j} ~\nabla_x c_{\rm e} - \mathbf{i}_{\rm e} \frac{t^0_+}{F}\right) + J_j,\label{eqn:elSpeciesConservation}
\end{equation}
where $\epsilon_{{\rm e},j}$ is the domain-specific electrolyte volume fraction, $t$ is time, $D^{\rm eff}_{{\rm e},j}$ is the domain-specific effective diffusion coefficient accounting for porous electrode effects~\cite{SB17}, $\mathbf{i}_{\rm e}$ is the electrolyte current density, $t^0_{+}$ is the Li-ion transference number, and $F$ is Faraday's constant.  The subscript $x$ indicates that the operator is in the $x$-direction (i.e., the through-plane direction). The Li-ion sink term due to electrochemical reactions $J_j$ can be represented as 
\begin{equation}
    J_j = \frac{3\epsilon_{{\rm AM},j}}{R_{{\rm s},j}}\dot{s}_j,
\end{equation}
where $\epsilon_{{\rm AM},j}$ is the active material volume fraction, $R_{{\rm s},j}$ is the representative active material particle radius, and $\dot{s}_{j}$ is the production rate of Li-ions due to electrochemical reactions. To avoid confusion, $J_j$ is defined differently between Part~I and Part~II.  This difference arises because the electrochemical reaction is heterogeneously distributed across the electrode in the P2D model as opposed to being homogeneously distributed in the SPM.  The current density in the electrolyte phase $\mathbf{i}_{\rm e}$ for a dual-salt concentrated system can be expressed as
\begin{equation}
	\mathbf{i}_{\rm e} = -{\kappa^{\rm eff}_{{\rm e},j}}\nabla_x\phi_{\rm e}- \kappa^{\rm eff}_{{\rm D},j} \nabla_x \ln c_{\rm e},
\end{equation}
where $\kappa^{\rm eff}_{{\rm e},j}$ is the effective electrolyte conductivity, $\kappa^{\rm eff}_{{\rm D},j}$ is the effective diffusive conductivity.  The effective electrolyte properties account for porous media effects.  The effective properties are related to the bulk-phase properties through the Bruggeman relation, which can be expressed as
\begin{align}
	D^{\rm eff}_{{\rm e},j} &= D_{\rm e} \epsilon^{p_{{\rm e},j}}_{{\rm e},j},\\
	\kappa^{\rm eff}_{{\rm e},j} &= \kappa_{\rm e} \epsilon^{p_{{\rm e},j}}_{{\rm e},j},\label{eqn:elEffectiveConductivity}\\
	\kappa^{\rm eff}_{{\rm D},j} &= \frac{2RT}{F} \kappa^{\rm eff}_{{\rm e},j}\left(t^0_{+}-1\right)\left(1+\frac{\partial \ln f_{\pm}}{\partial\ln\left(c_{\rm e}\right)}\right),\label{eqn:elDiffusionConductivity}
\end{align}
where $p_{{\rm e},j}$ is the Bruggeman exponent in the electrolyte phase, $D_{\rm e}$ is the bulk-phase Li-ion diffusion coefficient, $k_{\rm e}$ is the bulk-phase ionic conductivity, $R$ is the universal gas constant, $T$ is temperature, and the final parenthetical term in Eq.~\ref{eqn:elDiffusionConductivity} is referred to as the thermodynamic factor.  Equation~\ref{eqn:elDiffusionConductivity} includes more than just the Bruggeman relation and is formulated to satisfy concentration solution theory for two charged species in a solution~\cite{CK10,NT04,FDN94}.  The Li-ion species conservation equation (Eq.~\ref{eqn:elSpeciesConservation}) is a parabolic, partial differential equation that has no-flux conditions at the anode and cathode current-collector and requires an initial salt concentration profile (typically assumed to be uniform).  

The electrolyte potential is resolved through conservation of charge as
\begin{equation}
	\nabla_x\cdot \mathbf{i}_{\rm e}= J_j F,\label{eqn:elChargeConservation}
\end{equation}
Conservation of charge in the electrolyte (Eq.~\ref{eqn:elChargeConservation}) is expressed here as an \emph{algebraic constraint} equation (i.e., there is no time-derivative). In other words, the governing equation only requires the specified no-flux condition at either electrode current-collector.  An initial condition is not necessarily specified \textit{a priori}, which requires special handling in the PINN P2D surrogate model (Section~\ref{sec:p2dSurr}).

The solid-phase charge conservation is assumed to be well represented using Ohm's law and can be expressed as
\begin{align}
    \nabla_x\cdot\left(\sigma^{\rm eff}_{{\rm s},j}~ \nabla_x \phi_{{\rm s},j}\right) = -J_j F,\label{eqn:edConservationCharge}
\end{align}
where $\sigma^{\rm eff}_{{\rm s},j}$ is the effective solid-phase electrical conductivity. The effective conductivity can be specified using a Bruggeman expression
	\begin{equation}
		\sigma^{\rm eff}_{{\rm s},j} = \sigma_{{\rm s},j} \epsilon_{{\rm s},j}^{p_{{\rm s},j}},
	\end{equation}
where $\sigma_{{\rm s},j}$ is the bulk solid-phase electrical conductivity, $\epsilon_{{\rm s},j}$ is the solid-phase volume fraction, and $p_{{\rm s},j}$ is the solid-phase Bruggeman coefficient.  The solid-phase conservation equations are only resolved in the anode and cathode domains.  The boundary conditions for this equation include no-flux conditions at the anode/separator and cathode/separator interfaces. The potential of the one of the electrode current-collectors is set to a reference value (e.g., $\phi_{\rm s,an}|_{x=0} = 0$) and the other electrode current-collector boundary condition is either set to a Dirichlet boundary condition (in the case of a specified voltage) or a Neumann boundary condition (in the case of current demand).  For the present study, a Neumann condition is specified at the cathode current-collector as
\begin{equation}
    \mathbf{n}_x \cdot \left( \sigma^{\rm eff}_{{\rm s,ca}} \nabla_x \phi_{\rm s,ca}\right)\Bigg|_{x=L_{\rm an} + L_{\rm sep}+L_{\rm ca}} = \frac{I}{A},
\end{equation}
where $\mathbf{n}_x$ is the unit normal in the $x$-direction, $L_{\rm an}$ is the anode thickness, $L_{\rm sep}$ is the separator thickness, and $L_{\rm ca}$ is the cathode thickness, $I$ is the current demand (in Amps), and $A$ is the battery geometric area.  Like the governing equation for the liquid-phase potential $\phi_{\rm e}$ (Eq.~\ref{eqn:elChargeConservation}), the governing equation for the solid-phase potential $\phi_{{\rm s},j}$ is an \emph{algebraic constraint} equation, which requires special handling in the PINN.

Conservation of solid-phase Li is simulated in the secondary $r$ direction as 
	\begin{align}
		\frac{\partial c_{{\rm s},j}}{\partial t}= \frac{1}{r^2}\frac{\partial}{\partial r}\left(D_{{\rm s},j}r^2\frac{\partial{c_{{\rm s},j}}}{{\partial r}}\right),\label{eqn:edSpeciesConservation}
	\end{align}
where $D_{{\rm s},j}$ is the solid-phase Li diffusion coefficient.  Like the SPM (Part~I), this parabolic governing equation has a no-flux condition at the particle center due to symmetry, and has a Li surface flux due to electrochemical reactions which can be expressed as
\begin{equation}
    \left(D_{{\rm s},j} \frac{\partial c_{{\rm s},j}}{\partial r}\right)_{r=R_{{\rm s},j}} = -\dot{s}_j.\label{eqn:ParticleBC}
\end{equation}

The net production of lithium ions due to charge-transfer reactions $\dot{s}_{j}$ couples the electrolyte/electrode equation set.  The charge-transfer reaction at the electrode/electrolyte surface is commonly expressed using a Butler--Volmer expression as
	\begin{align}
		\dot{s}_{j} = \frac{i_{0,j}}{F}\left[\exp\left(\frac{\alpha_{{\rm a},j} F \eta_j}{RT}\right)-\exp\left(\frac{(\alpha_{{\rm a},j}-1) F \eta_j}{RT}\right)\right],
	\end{align}
where $\alpha_{a,j}$ is the anodic symmetry factor (assumed to be 0.5 in both the anode and cathode domains), $i_{0,j}$ is the exchange current density, and $\eta_j$ is the kinetic overpotential. The exchange current density $i_{0,j}$ can be expressed as
\begin{equation}
    i_{0,j} = i^0_{0,j} c_{\rm e}^{\alpha_{{\rm a},j}} \left(c_{{\rm s,max},j} - c_{{\rm s},j}|_{r=R_{{\rm s},j}}\right)^{\alpha_{{\rm a},j}} \left(c_{{\rm s},j}|_{r=R_{{\rm s},j}}\right)^{(1-\alpha_{{\rm a},j})}.
\end{equation}
The kinetic overpotential $\eta_j$ can be expressed as
\begin{equation}
    \eta_j = \phi_{{\rm s},j} - \phi_{\rm e} - U_{{\rm OCP},j}\left(c_{{\rm s},j}|_{r = R_{{\rm s},j} }\right),
\end{equation}
where $U_{{\rm OCP},j}$ is the electrode open-circuit potential evaluated using the solid-phase surface concentration.

The spatiotemporal domain considered for P2D equations is the same as in Part~I for the time variable $t$ ([0, 1350s]) and the radial location $r$ ([0, 4$\mu \rm{m}$] in the anode and [0, 1.8$\mu \rm{m}$] in the cathode). In the transversal direction, the $x$-domain is defined as [0, 44$\mu \rm{m}$] for the anode,  [44$\mu \rm{m}$, 64$\mu \rm{m}$] for the separator, [64$\mu \rm{m}$, 106$\mu \rm{m}$] for the cathode.

%% file: part2_termvolt.tex
\begin{figure*}[th!]
    \centering
    \includegraphics[width=6.376in]{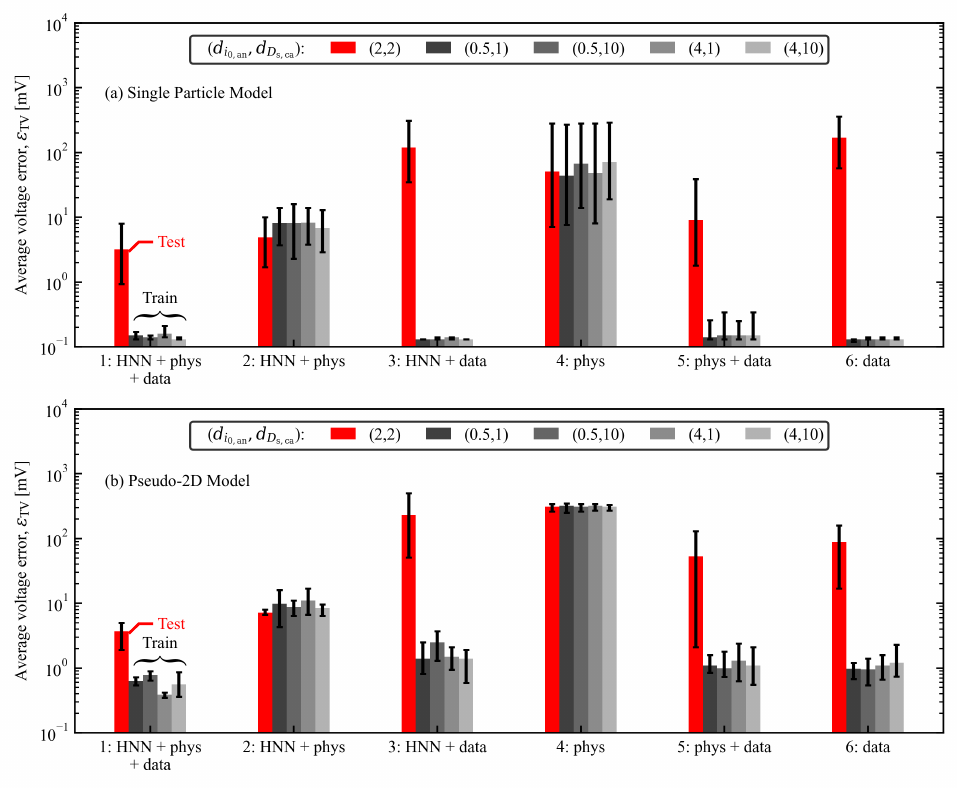}
    \caption{\revtwotwo{Average terminal voltage error $\varepsilon_{\rm TV}$ (bar height) over the parameter range considered for the exchange current density in the anode $i_{\rm 0,an}$ and the \ce{Li} diffusivity in the cathode $D_{\rm s,ca}$. The red bar shows the accuracy result for a parameter set not included in the data set. The black bars show the accuracy for the parameter sets included in the data set. The error bar denotes the 95\% percentile variability observed for all the realizations.}}
    \label{fig:par_var_tv}
\end{figure*}

\revtwotwo{Throughout the analysis in Sec.~\ref{sec:parametericPinn}, the relative error included contributions from all state variables. This is a useful metric to show how well the PINN surrogate captures the dynamics for all of the state variables. For the purpose of diagnostics, however, only the cathode current collector voltage needs to be predicted. Therefore, it is also informative to evaluate the terminal voltage error, i.e., the error in the prediction of the cathode current collector voltage. Additionally, it is useful to compare the terminal voltage error for the P2D and the SPM (Eq.~\ref{eq:errP2D}). As the relative error metric (shown in the main manuscripts) depend on the number of state variables, there is an inherent difference between the P2D model and the SPM that makes comparisons difficult. However, it is reasonable to compare the terminal voltage error metric.  The terminal voltage error is defined similarly to Part I~\cite{hassanaly2023pinn1} as }
\begin{equation}
    \revtwotwo{\varepsilon_{\rm TV} = \frac{1}{N_{\xi}} \sum_{i \in [1,N_{\xi}]} \left| \phi_{\rm s, c, CC, PINN} - \phi_{\rm s, c, CC, PDE} \right|},
\end{equation}
\revtwotwo{where $\phi_{\rm s, c, CC, PDE}$ is the potential at the cathode current collector obtained from finite difference at the point $i$, $\phi_{\rm s, c, CC, PINN}$ is the predicted potential at the cathode current collector at the point $i$, and $N_{\xi}$ is the number of points over which the error is computed. Figure~\ref{fig:par_var_tv} shows the terminal voltage errors for both the SPM and P2D model for the test and training conditions (similar to Fig.~\ref{fig:par_var}).}

\revtwotwo{Overall, the trends observed in Sec.~\ref{sec:parametericPinn} are also reflected in the terminal voltage error metric. The errors incurred on the P2D model are however closer to the SPM which suggests that the main contributions of the relative error metric for the P2D model were not due to $\phi_{\rm s, ca}$. However, the errors for case 1 (\textit{HNN+phys+data}) are on average higher than for the SPM, including where data is used.}

%% file: SPM_Brute_revised_highlights.tex
\begin{figure}[ht!]
    \centering
    \includegraphics[width=2.403in]{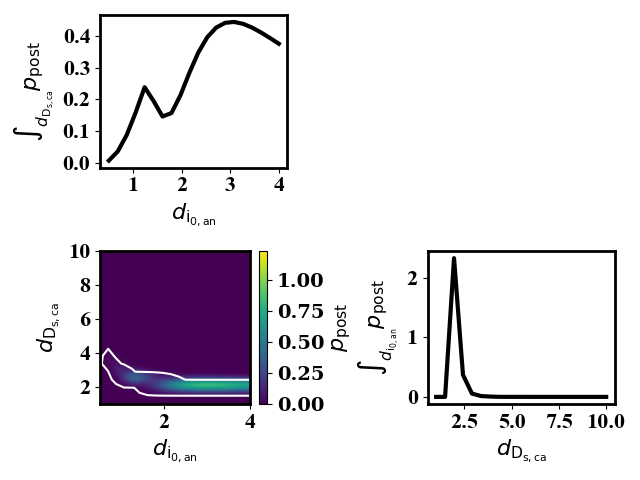}
    \caption{Posterior probability $p_{\rm post}$ obtained from integration of 400 realizations of the SPM PDE (bottom left). Marginal posterior PDF with respect to $d_{\rm i_{0,an}}$ (top left) and $d_{\rm D_{s,ca}}$ (bottom left).}
    \label{fig:brute_post}
\end{figure} 

In this section\revtwothree{,} the posterior PDF obtained with the PINN surrogate of the SPM is validated. Instead of performing the Bayesian calibration with an MCMC procedure, the posterior is directly computed over the entire parameter space. This is possible for the specific case tackled here given that the inferred parameter space is only two\revtwothree{-}dimensional. The product of $p_{\rm prior} (\boldsymbol{p}) p_{\rm like} (\boldsymbol{d} | \boldsymbol{p})$ is evaluated for the noisy case by integrating the PDE of the SPM over 400 points that uniformly span the parameter space. Given the higher cost of the P2D model integration, this strategy is only doable for the SPM. The uncertainty in the likelihood function is kept equal to $5.36~\rm{mV}$ to be consistent with Sec.~\ref{sec:cal}. The actual posterior $p_{\rm post} (\boldsymbol{p}|\boldsymbol{d})$ is reconstructed by enforcing the realizability constraint ($\int p_{\rm post} (\boldsymbol{p}|\boldsymbol{d}) d\boldsymbol{p} = 1$).

The posteriors obtained \revtwothree{are} shown in Fig.~\ref{fig:brute_post}. It shows distinctive features that were also captured by the PINN surrogate (Fig.~\ref{fig:spm_p2d_cal} d-f). First, $d_{D_{\rm s, ca}}$ can be identified more easily than $d_{i_{\rm 0, an}}$. Second, the available measurements are insufficient to identify the value of $d_{i_{\rm 0, an}}$. On top of being wide, the support of the marginal posterior of $d_{i_{\rm 0, an}}$ exhibits a bimodal structure also seen in Fig.~\ref{fig:spm_p2d_cal}e. Finally, the curved shape of the joint posterior support (shown with the white contour in Fig.~\ref{fig:brute_post}) also resembles the contour of the joint posterior shown in Fig.~\ref{fig:spm_p2d_cal}e.